%% file: main.tex
\documentclass[11pt]{article}

\usepackage{acl}

\usepackage{times}
\usepackage{latexsym}
\usepackage[T1]{fontenc}
\usepackage[utf8]{inputenc}
\usepackage{microtype}
\usepackage{inconsolata}

\input{preamble}

\usepackage{graphicx}
\usepackage{booktabs}
\usepackage{multirow}
\usepackage{dsfont}
\usepackage{amsmath}
\usepackage{amsfonts}
\usepackage{tcolorbox}
\usepackage{float} 

\title{Look Where It Matters: High-Resolution Crops Retrieval for Efficient VLMs}

\author{
  \textbf{Nimrod Shabtay\textsuperscript{1,2}\thanks{Equal contribution.}},
  \textbf{Moshe Kimhi\textsuperscript{3}\footnotemark[1]},
  \textbf{Artem Spector\textsuperscript{1}},
  \textbf{Sivan Haray\textsuperscript{1}},
  \textbf{Ehud Rivlin\textsuperscript{3}},
\\
  \textbf{Chaim Baskin\textsuperscript{4}},
  \textbf{Raja Giryes\textsuperscript{2}},
  \textbf{Eli Schwartz\textsuperscript{1}}
\\
  \textsuperscript{1}IBM Research,
  \textsuperscript{2}Tel-Aviv University,
  \textsuperscript{3}Technion,
  \textsuperscript{4}Ben-Gurion University
}

\begin{document}
\maketitle

\begin{abstract}
Vision-language models (VLMs) typically process images at a native high-resolution, forcing a trade-off between accuracy and computational efficiency: high-resolution inputs capture fine details but incur significant computational costs, while low-resolution inputs offer efficiency at the cost of potentially missing critical visual information, like small text. We present \method{}, a spatial-on-demand framework that resolves this accuracy--efficiency trade-off by operating on a low-resolution global view and using tool-calling to retrieve only high-resolution segments needed for a given query.
We construct supervised data automatically: a judge compares low- vs.\ high-resolution answers to label whether cropping is needed, and a grounding model localizes the evidence for the correct answer, which we map to a discrete crop set to form multi-turn tool-use trajectories.
We train our framework with cold-start SFT followed by multi-turn GRPO with a composite reward that combines semantic answer correctness with explicit crop-cost penalties.
Extensive evaluation across six benchmarks demonstrates that \method{} matches full-resolution accuracy while retaining only 36\% of visual tokens.
\newline \url{https://nimrodshabtay.github.io/AwaRes}
\end{abstract}

\begin{figure*}[t]
\centering
\includegraphics[width=0.85\linewidth]{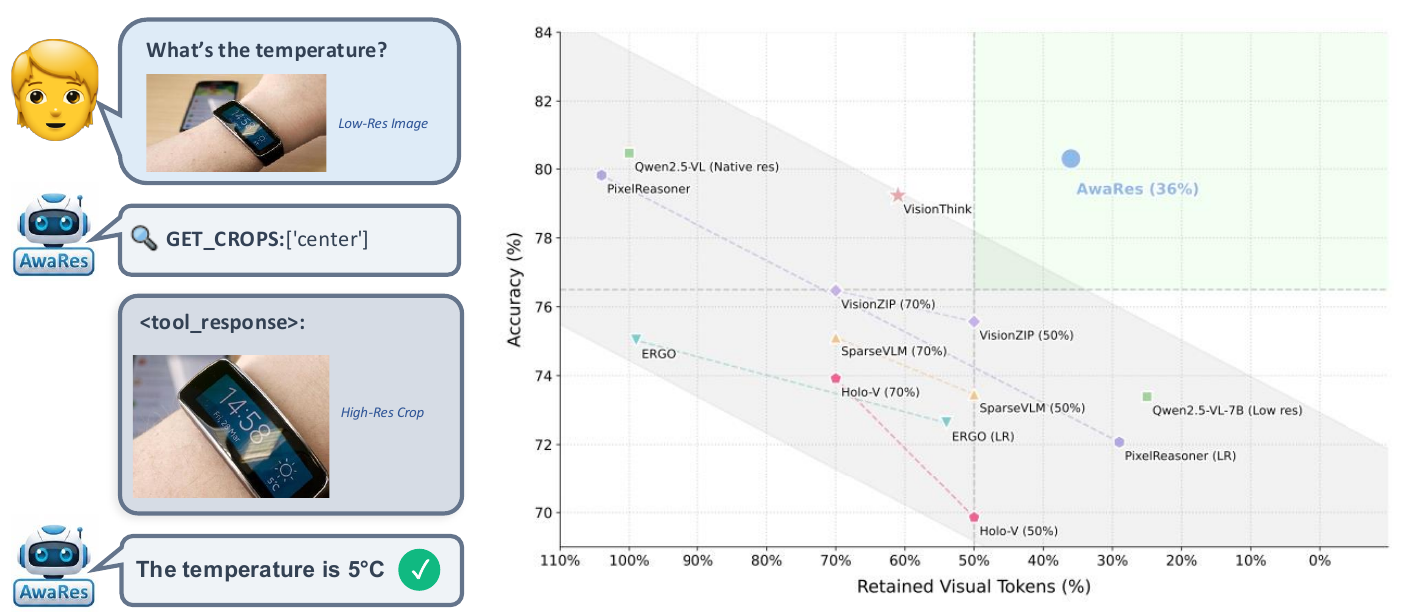}
\caption{\method{} overview. Left: Given a low-resolution image, \method{} uses tool-calling to request only the high-resolution crops needed to answer the query. Right: Accuracy vs. retained visual tokens across six benchmarks. \method{} performs similarly to native high-resolution (80.3\%) while using only 36\% of the visual tokens.}
\label{fig:teaser}
\end{figure*}

\section{Introduction}
\label{sec:intro}
Vision–language models (VLMs) increasingly rely on high-resolution visual inputs to solve detail-sensitive tasks such as document question answering, chart understanding, and understanding semantics and text in dense natural images. However, high resolution is expensive: the number of visual tokens grows rapidly with image resolution, making high-resolution inference a major bottleneck in practice.
Existing approaches to reduce this cost largely fall into several camps.
First, \emph{token pruning} methods selectively discard visual tokens to reduce computation~\cite{fastv,Pyramiddrop,VisionZip,SparseVLM,HoloV}. While effective in principle, they often introduce irregular token patterns and dynamic sequence lengths that can be difficult to translate into end-to-end serving speedups in common inference stacks, such as vLLM~\cite{vllm}, where efficiency is tied to predictable sequence length.
Second, \emph{resolution escalation} methods~\cite{VisionThink,CARES} learn when to request a higher-resolution view, but typically treat the decision as binary: if more details are needed, the entire high-resolution image is retrieved, wasting computation on regions irrelevant to the question.
A key observation is that the demand for high fidelity is usually \emph{spatially sparse}, as can be seen in Fig.~\ref{fig:data_sampley}.
Many questions require fine detail in only a small portion of the image: a single value on a chart axis, a specific cell in a table, or a tiny object in the corner of an image.
Recent works have begun to exploit this property by training VLMs to selectively zoom into task-relevant regions~\cite{lee2025ergo, wang2025pixel, DeepEyes}. However, these methods rely on an iterative reasoning process-where latency is dominated by long reasoning trace rather than the visual token budget per round; reducing input resolution alone yields only marginal speedups for these methods, since each additional reasoning step, not pixel count, is the primary cost. 
We advocate that answering the question of \emph{where} to look matters as much as \emph{whether} to look—and that it can be achieved without heavyweight reasoning machinery.

We propose \method{}, a framework that exploits this spatial sparsity via a simple tool-calling interface that targets \textbf{high-resolution crop acquisition}. \method{} processes a low-resolution global view by default, and when additional detail is required, it invokes a tool-call that requests only specific high-resolution sub-regions, and then answers conditioned on both.
This multi-turn structure is naturally compatible with KV-caching: computation from the initial low-resolution turn is reused and extended in the crop turn without architectural changes, making \method{} practical for deployment.
We train \method{} to learn a single \emph{coupled-decision policy (CDP)} that jointly decides (i) \emph{whether} additional resolution is needed and (ii) \emph{where} to acquire it by selecting a subset of crops.
Crucially, these decisions are \emph{fused} into the model's first-turn action: either answer directly, or emit a structured crop request that simultaneously signals escalation and specifies the target regions.
For the cold-start phase, we construct the supervision automatically, without manual spatial annotations, by (i) identifying examples where the low-resolution model's outputs are insufficient using an LLM-as-a-Judge (LaaJ), and (ii) localizing the evidence for the correct answer using a grounding model to produce target crops.
We evaluate \method{} on six benchmarks spanning document understanding and general visual QA.
Across these tasks, \method{} also resolves the same whether-and-where decision much more efficiently, avoiding the reasoning overhead and achieving on average a $5.7\times$ latency reduction over reasoning-based methods, and a $4.4\times$ reduction over the resolution-escalation method \cite{VisionThink}, while retaining only 36\% of visual tokens (RTR) and matching full-resolution accuracy ($80.3\%$ vs. $80.46\%$).
Our contributions are listed as follows:
(1) We introduce a spatial-on-demand inference framework that acquires targeted high-resolution crops through tool-calling, enabling system-friendly multi-turn KV-cache reuse.
(2) We propose an automatic data curation pipeline that produces multi-turn tool-use trajectories without manual annotations.
(3) We refine crop usage with multi-turn GRPO using an objective that penalizes unnecessary crop acquisition while discouraging missed crop requests when detail is required.

\section{Related Work}
\label{sec:related_work}
Several strategies have emerged to reduce the number of visual tokens in Vision Language Models.

One line of research focuses on dynamic token pruning. Methods such as FastV \cite{fastv}, HoloV \cite{HoloV}, PyramidDrop \cite{Pyramiddrop}, FitPrune \cite{fitprune}, TopV \cite{TopV}, SparseVILA \cite{SparseVILA}, IVTP \cite{ivtp}, LLaVolta \cite{llavolta}, and SAINT \cite{saint} discard uninformative tokens within the LLM layers. Alternatively, others prune tokens directly after the vision encoder \cite{VisionZip, FastVLM, SparseVLM,kimhi2025wave}. While effective, these approaches generally commit to a fixed token budget regardless of the specific instance's complexity. In contrast, our method is fully adaptive, allowing simple images to be processed at minimal cost while allocating more resources only when and where the query demands fine-grained perception.

A second line of work explores resolution selection. CARES \cite{CARES} uses an external lightweight model to predict the optimal input resolution, while CROP \cite{CROP} identifies contextual regions of interest via an auxiliary module. Because these methods rely on external components, they require hosting multiple models simultaneously. Our approach, by contrast, natively enables the VLM itself to determine where additional detail is needed.

A third line of work trains VLMs to actively escalate resolution or selectively zoom into task-relevant regions. VisionThink \cite{VisionThink} issues a tool call to retrieve the entire high-resolution image when the low-resolution view is insufficient, wasting computation on irrelevant areas. Alternatively, ZoomEye \cite{ZoomEye}, DeepEyes \cite{DeepEyes}, ERGO \cite{lee2025ergo}, and Pixel Reasoner \cite{wang2025pixel} achieve spatial selectivity through multiple inference passes or explicit visual operations embedded within long Chain-of-Thought (CoT) reasoning. While effective at improving accuracy, these approaches share a critical bottleneck: their reliance on extended autoregressive reasoning traces or massive full-image token counts severely degrades generation speed. Consequently, they fail to translate into real-world deployment latency benefits. Our method sidesteps this overhead entirely. By emitting a single structured tool call in the first turn to request only targeted high-resolution crops, \method{} bypasses intermediate reasoning steps and full-image overhead, achieving spatial-on-demand selectivity at a fraction of the inference cost.

\section{Method}
\label{sec:method}
\method{} implements \emph{spatial-on-demand} perception via a simple multi-turn interaction: the model first observes a low-resolution global view, and only if needed issues a tool call to retrieve a set of high-resolution crops (Fig.~\ref{fig:teaser}).
We first formalize this interaction protocol (\S\ref{sec:problem_setup}), then describe how we automatically curate supervision for the CDP, namely \emph{whether} additional resolution is needed and \emph{where} it matters (Fig.~\ref{fig:data_pipeline}; \S\ref{sec:data_curation}).
Finally, we train in two stages: (i) a \emph{cold-start} supervised fine-tuning (SFT) stage that teaches the tool protocol and yields a supervised \emph{reference policy} $\pi_{\text{ref}}$ (\S\ref{sec:sft}); and (ii) multi-turn GRPO initialized from $\pi_{\text{ref}}$ and regularized toward it via a KL penalty (\S\ref{sec:grpo}), explicitly optimizing the accuracy--efficiency trade-off.

\subsection{Problem setup}
\label{sec:problem_setup}
Given an image--question--answer triple $(I, q, a^\star)$, the model is first shown a low-resolution view $I_{\text{low}}$ (obtained by downsampling $I$) together with the question $q$.
The model then chooses between two actions:

\noindent\textbf{(i) Direct answer:} an answer $\hat{a}$ based on $(q, I_{\text{low}})$.

\noindent\textbf{(ii) Crop request + answer:} Emits a tool call that requests a \emph{subset} of crops from a predefined candidate set, $\mathcal{C}_{\text{req}} \subseteq \mathcal{C}$.
The tool returns the corresponding high-resolution crop images $\{I^{\text{high}}_{c}\}_{c\in \mathcal{C}_{\text{req}}}$, which are appended to the dialogue context, and the model produces the final answer $\hat{a}$ conditioned on the full multi-turn history.

\noindent\textbf{A fused coupled-decision policy:}
We parameterize a single policy over high resolution request, and localized crop selection:
\begin{equation}
    \pi_\theta(C \mid q, I_{\text{low}}), \qquad C \subseteq \mathcal{C},
\end{equation}
where $C=\emptyset$ corresponds to \emph{no tool call} (answer directly) and $C\neq\emptyset$ corresponds to \emph{escalation with localization}.
Under this view, ``when to crop'' is the marginal event $\mathds{1}[C\neq\emptyset]$, while ``where to crop'' is the conditional distribution over $C$ given $C\neq\emptyset$.
The two are inherently coupled: the value of escalating depends on \emph{which} regions will be retrieved, since inaccurate localization can waste compute without improving answer correctness.

\noindent This interface targets efficiency by restricting high-resolution perception to a small number of structured regions, while preserving the low-resolution global context throughout the interaction (See Fig.~\ref{fig:teaser} for a conversation example).

\subsection{Automatic supervision for crop requests}
\label{sec:data_curation}
\begin{figure}
    \centering
    \includegraphics[width=\linewidth]{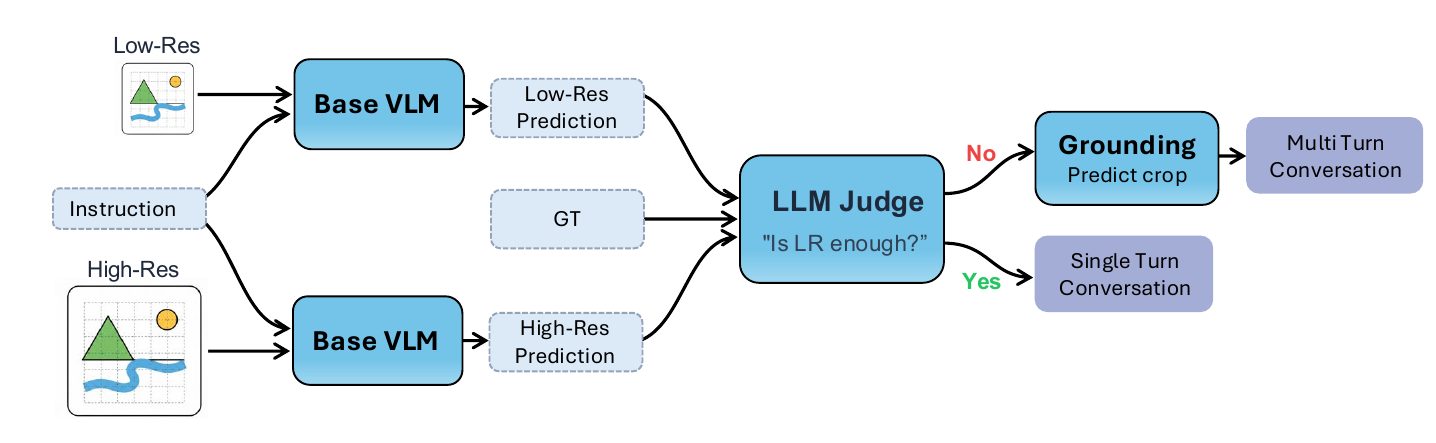}
    \caption{\textbf{Overview of the automatic supervision pipeline.} Each sample is processed at two resolutions; an LLM judge determines resolution sufficiency by comparing predictions to ground truth. Sufficient cases yield single-turn conversations, while insufficient cases are routed to a grounding model for crop localization, producing multi-turn trajectories with tool-calling.}
    \label{fig:data_pipeline}
\end{figure}

A key challenge is to supervise two coupled decisions: whether the low-resolution view is insufficient, and \textbf{where} to crop when additional detail is needed.
We generate this supervision to initiate the model to learn a reference policy $\pi_{\theta_{ref}}$ in an automatic fashion using the three-stage pipeline (Illustrated in Fig.~\ref{fig:data_pipeline}):

\noindent\textbf{Stage 1: resolution-sufficiency labeling (when to crop).}
For each example $(I, q, a^\star)$ we utilize a base VLM $T$ on both the low-resolution and full-resolution inputs:
\begin{equation}
\hat{a}_{\text{low}} = T(q, I_{\text{low}}),
\qquad
\hat{a}_{\text{full}} = T(q, I).
\label{eq:low_full_answers}
\end{equation}

Because $\hat{a}_{\text{low}}$ and $\hat{a}_{\text{full}}$ may differ in form though semantically correct, direct string matching (exact match) with $a^\star$ is unreliable.
Instead, we use an LaaJ (LLaMA-3.3-70B~\cite{llama3}) to compare both predictions to the ground truth $a^\star$.
If it judges $\hat{a}_{\text{low}}$ as correct (or ties it with $\hat{a}_{\text{full}}$), we label the example as no crop needed \texttt{LR}; otherwise we label it as \texttt{HR}.

\noindent\textbf{Stage 2: crop target construction (where to crop).}
For examples labeled \texttt{HR}, we identify the region that contains the visual evidence needed to answer $(q, a^\star)$.
We prompt a grounding model $G$ (namely, Qwen3-VL-A235B-A22B~\cite{qwen3_vl}) to localize the evidence and return a bounding box
$b = (x_1, y_1, x_2, y_2)$ in the coordinate system of the original image.

\begin{figure*}[t]
    \centering
    \includegraphics[width=0.8\linewidth]{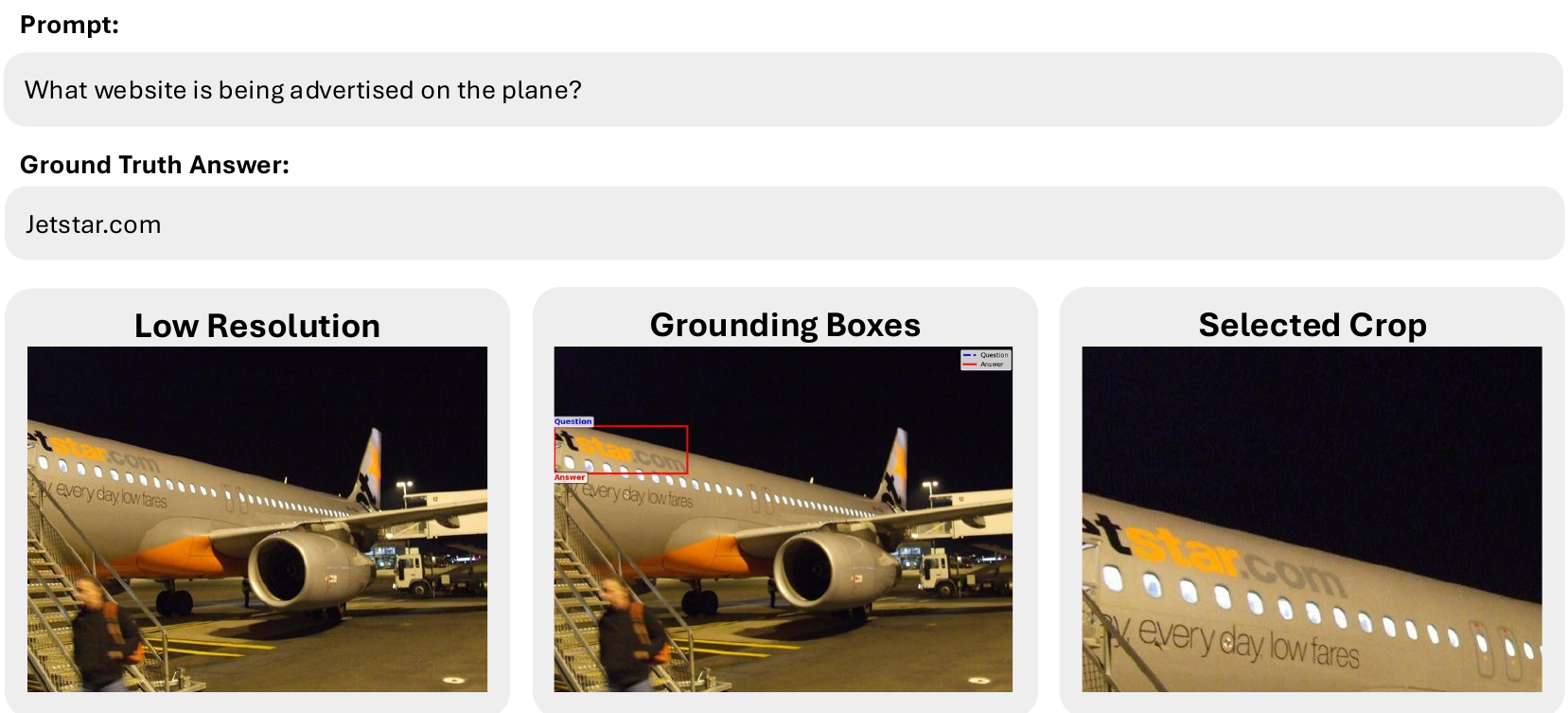}\\
    \includegraphics[width=0.8\linewidth]{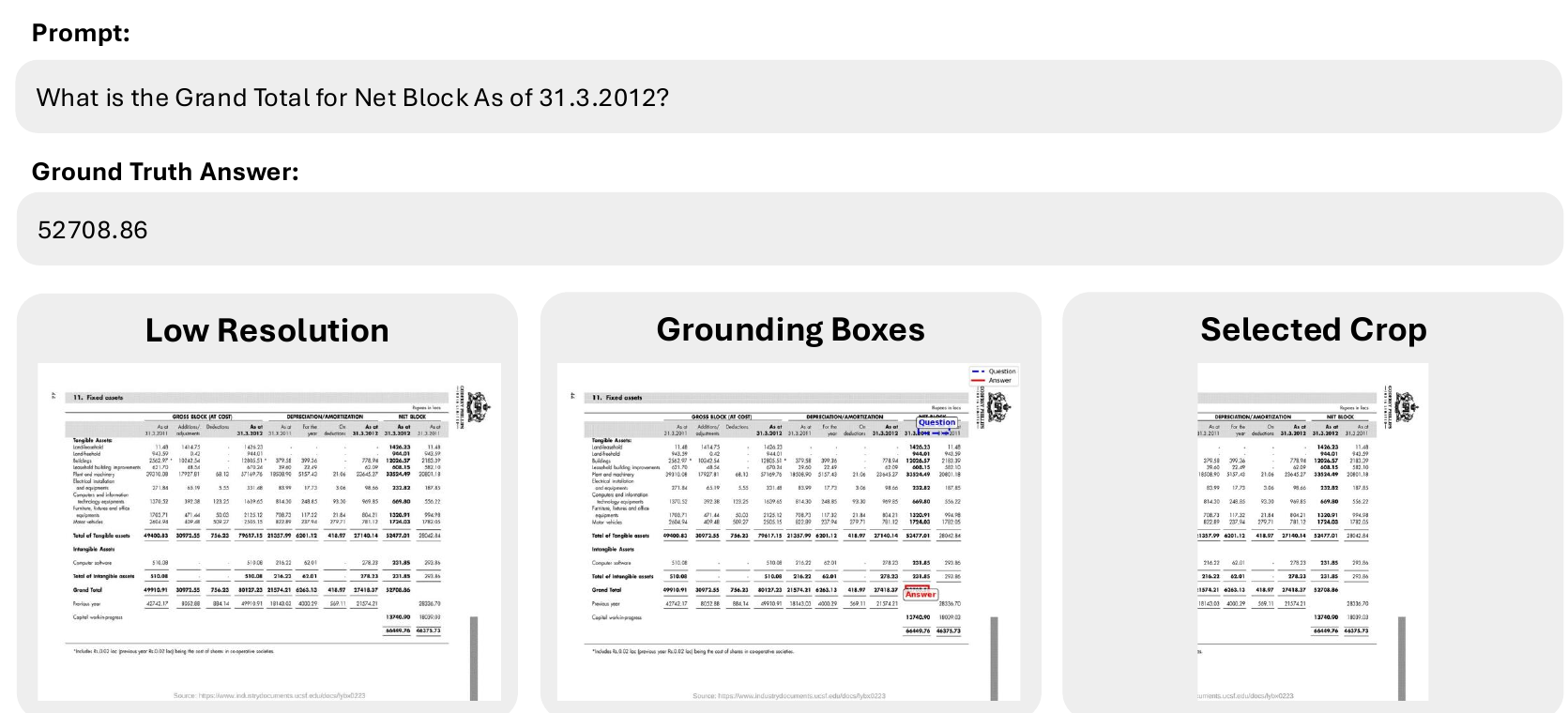}
    \caption{\textbf{Crop annotation example.} Left: low-resolution input where text is illegible. Middle: grounding model $G$ predicted bounding box localizing the answer region. Right: selected high-resolution crop enabling correct response (best viewed when zoomed in).}
    
    \label{fig:data_sampley}
\end{figure*}

We then map $b$ to our discrete crop candidate set $\mathcal{C}$, which includes four quadrants, a center crop, four merged half-image regions (top/bottom/left/right), and a full-image.
We define the target crop subset as
\begin{equation}
\mathcal{C}^\star = \{\, c \in \mathcal{C} \;|\; \operatorname{IoU}(b, c) \ge \tau \,\},
\label{eq:crop_mapping}
\end{equation}
where $\tau = 0.5$ is the IoU threshold.
Fig.~\ref{fig:data_sampley} shows a representative example, and Fig.~\ref{fig:crop_flow_between_phases} (left side) summarizes the empirical distribution of selected crops in the curated training set.

\noindent\textbf{Stage 3: supervised tool-use trajectories.}
The procedure above yields two types of training transcripts:

\noindent\textbf{Direct-answer trajectories (\texttt{LR}).}
The model observes $(q, I_{\text{low}})$ and is supervised to output $a^\star$ in a single turn.

\noindent\textbf{Tool-call-then-answer trajectories (\texttt{HR}).}
In the first turn, the model issue a tool call selecting $\mathcal{C}^\star$.
After the tool returns $\{I^{\text{high}}_{c}\}_{c\in \mathcal{C}^\star}$, the model is trained to produces $a^\star$ in a second turn conditioned on both the low-resolution and the retrieved high-resolution crops.

This curation pipeline produces multi-turn tool-use supervision at scale in order to learn an initial reference policy $\pi_{\theta_{ref}}$, while keeping the crop interface structured and deployment-friendly (Fig.~\ref{fig:data_pipeline}).
We provide additional details in the Appendix.

\subsection{Cold-start (SFT)}
\label{sec:sft}
We cold-start our crop-request policy by supervised fine-tuning (SFT) on the mixture of direct-answer and tool-call-then-answer trajectories produced in \S\ref{sec:data_curation}.

This stage serves two purposes: (i) teach the model to follow the multi-turn tool-calling protocol and learn the coupled decisions (\emph{whether} additional detail is needed and \emph{where} it matters), and (ii) produce a strong supervised \emph{reference policy} $\pi_{\text{ref}}$ that we later use for KL-regularized GRPO (\S\ref{sec:grpo}).

Let $y_{1:T}$ denote the assistant tokens in a supervised transcript, and let $h_t$ be the dialogue history at step $t$ (including $(q, I_{\text{low}})$, any previously generated tokens, and tool outputs if a crop request occurred).
We minimize a weighted negative log-likelihood:

\begin{equation}
\mathcal{L}_{\text{SFT}}(\theta) = -\sum_{t=1}^{T} w_t \log \pi_\theta(y_t \mid h_t)
\label{eq:sft_weighted_clean}
\end{equation}

The tool-call turn, despite having small numbers of tokens, fully specifies the CDP action and carry disproportionate control over both efficiency and downstream answer quality.
Upweighting this turn therefore directly stabilizes learning of the \emph{fused} first-turn decision.
After SFT, we freeze the resulting model as the reference policy $\pi_{\text{ref}}$ and initialize GRPO from it.

\subsection{Multi-turn GRPO}
\label{sec:grpo}
After the cold-start SFT stage, the model reliably follows the tool protocol but tends to \emph{over-request} crops even when $I_{\text{low}}$ is sufficient.
We therefore apply Group Relative Policy Optimization (GRPO) on full multi-turn interactions to explicitly optimize the accuracy--efficiency trade-off.

We denote by $\pi_{\text{ref}}$ the \emph{frozen} SFT policy $\pi_{\theta_{\text{ref}}}$ obtained from the SFT.
GRPO is initialized from $\pi_{\text{ref}}$ and uses a KL penalty to keep $\pi_\theta$ close to $\pi_{\text{ref}}$ while improving tool usage.

\noindent\textbf{Rollouts and trajectories.}
Given an input prompt $x=(q, I_{\text{low}})$, the policy $\pi_\theta$ generates first turn that may include a crop tool call.
The requested crops are appended to the dialogue context, and generation continues until a final answer $\hat{a}$ is produced.
We treat only assistant tokens as actions; tool outputs are treated as observations.
Thus, each rollout yields a multi-turn trajectory $\tau$ consisting of assistant actions interleaved with tool observations, ending with $\hat{a}$.

Unlike supervised training with dense per-token loss, GRPO enables optimization with task-specific rewards that directly target improved tool usage.

\subsubsection{Reward design.}
We assign a single scalar reward to the completed trajectory $\tau$, composed from two components:

\begin{equation}
\label{eq:reward}
    R(\tau) = w_a \cdot R_{\text{ans}}(\hat{a}, a^\star) - C_{\text{tool}}(C, y),
\end{equation}

\noindent\textbf{Answer reward} ($R_{\text{ans}}(\hat{a}, a^\star)$): measures semantic correctness using the cosine similarity between sentence-transformer embeddings of $\hat{a}$ and $a^\star$, where $w_a$ scales the answer reward relative to the tool cost.

\noindent\textbf{Tool-use cost:} Penalize tool usage with an asymmetric cost:
\begin{equation}
\resizebox{1.\linewidth}{!}{$\displaystyle
    C_{\text{tool}}(C, y) =
    \begin{cases}
        \alpha_{\text{miss}}
        & \text{if } y=\texttt{HR} \text{ and } C=\emptyset
        \quad \text{(missed tool-call)}\\[2pt]
        \alpha_{\text{use}} + \lambda \lVert C \rVert
        & \text{if } C\neq\emptyset
        \quad \text{(tool usage)}\\[2pt]
        0
        & \text{if } y=\texttt{LR} \text{ and } C=\emptyset,
    \end{cases}
    $}
\end{equation}
This asymmetry biases the policy toward \emph{recall} in tool invocation: missing a necessary crop request is penalized more heavily than making an unnecessary request.
When the tool is used, we additionally penalize the \emph{amount of high-resolution evidence requested} via $\lVert C \rVert$, defined as the total fraction of image area covered by the selected crops.
This encourages the policy to prefer smaller crops when they suffice.
Importantly, the cost depends on \emph{how much} is requested but remains agnostic to \emph{which} specific region is chosen, allowing the GRPO to explore alternative policies.

\subsubsection{GRPO optimization.}

For each prompt $x$, we sample a group of $G$ trajectories $\{\tau_1, \ldots, \tau_G\}$ from the current policy $\pi_\theta$. Each trajectory $\tau_i$ consists of a sequence of assistant tokens (actions) interleaved with tool observations, culminating in a final answer. We compute the advantage for each trajectory using the group-relative baseline:
\begin{equation}
    \hat{A}_i = \frac{R(\tau_i) - \mu_G}{\sigma_G + \epsilon},
\end{equation}
where $R(\tau_i)$ is the total reward for trajectory $\tau_i$, and $\mu_G$, $\sigma_G$ are the mean and standard deviation of rewards within the group.

We optimize a PPO-style clipped objective with KL regularization to the reference policy:
  \begin{equation}
  \resizebox{0.91\linewidth}{!}{$\displaystyle
      \begin{split}
      \mathcal{L}_{\text{GRPO}}(\theta) = \mathbb{E}_{x \sim \mathcal{D}} \Bigg[
      & \frac{1}{G} \sum_{i=1}^{G} \frac{1}{|\tau_i|} \sum_{t=1}^{|\tau_i|} \min \big( r_t^{(i)} \hat{A}_i, \, \text{clip}(r_t^{(i)}, 1-\epsilon, 1+\epsilon) \hat{A}_i
  \big) \\
      & - \beta \, D_{\text{KL}}(\pi_\theta \| \pi_{\text{ref}}) \Bigg]
      \end{split}
  $}
  \end{equation}
where $r_t^{(i)} = \frac{\pi_\theta(a_t^{(i)} | x, a_{<t}^{(i)})}{\pi_{\text{old}}(a_t^{(i)} | x, a_{<t}^{(i)})}$ is the importance sampling ratio, $\epsilon$ is the clipping threshold, and $\beta$ controls the strength of the KL divergence penalty against a reference policy $\pi_{\text{ref}}$.

As in PPO-style updates, $\pi_{\text{old}}$ denotes a snapshot of the policy before the current GRPO update step.
The KL term is computed over assistant-token distributions along the sampled trajectories, encouraging stable improvements over $\pi_{\text{ref}}$.

\subsection{Inference}
At test time, we follow the same interaction protocol used during training. The model receives $(q, I_{\text{low}})$ and either answers directly or emits a tool call selecting a crop subset $C \subseteq \mathcal{C}$. If a tool call occurs, the corresponding high-resolution crops are appended to the dialogue context while retaining $I_{\text{low}}$, and the model produces the final answer in a second turn.

This results in two possible inference paths: a single prefill pass for queries answerable from the low-resolution view, or two prefill passes when high-resolution detail is required. In the latter case, the model benefits from both the global context preserved in $I_{\text{low}}$ and the fine-grained detail in the requested crops. When a second turn is required the low-resolution view and the query are already in the KV cache saving on the required compute. Crucially, the decision of which path to take, and which regions to acquire - is made entirely by the learned policy, requiring no external heuristics or task-specific thresholds.

\section{Experimental Results}
\label{sec:results}
We evaluate \method{} on six benchmarks spanning document understanding and general visual QA, and compare against both fixed-budget token-pruning methods and adaptive resolution-escalation baselines.
We report (i) the dataset metric from \texttt{lmms-eval}~\cite{lmmseval} and (ii) an \emph{Retain Token Ratio} (RTR), defined as the fraction of visual tokens processed relative to the full-resolution baseline.
RTR directly reflects the model's first-turn \emph{coupled-decision policy} (answer directly vs.\ request crops), while accuracy reflects the quality of the full multi-turn interaction.
We first describe our evaluation protocol~\ref{subsec::eval_protocol}, evaluated datasets ~\ref{subsec::datasets} and implementation details~\ref{subsec::impl_details}. Then, we provide a detailed discussion of the main results~\ref{subsec::main_results} and conclude by extensive ablations~\ref{subsec::ablations}.

\subsection{Evaluation Protocol}
\label{subsec::eval_protocol}
All models are evaluated using \texttt{lmms-eval}~\cite{lmmseval}, and we report the per-dataset metrics provided by the framework.

\noindent\textbf{Retain Token Ratio (RTR):} We measure efficiency via \emph{visual token} usage, which dominates compute and KV-cache memory at high resolution. Let $T_i$ be the visual tokens a sample $i$ consumes across \emph{all} turns and $T_{\mathrm{full}}$ those of one full-resolution pass with the baseline model; then $\mathrm{RTR}_i = T_i/T_{\mathrm{full}}$ and $\mathrm{RTR} = \mathbb{E}_i[\mathrm{RTR}_i]$. Fixed-budget methods are configured to retain $50\%$ or $70\%$ of the full-resolution tokens; for adaptive methods we count the tokens actually consumed per sample and average over the dataset.

\noindent\textbf{Latency:} When reporting wall-clock time (Fig.~\ref{fig:performance_wallclock}), we measure end-to-end per-sample latency, including both turns for methods that invoke the crop tool.

\subsection{Datasets}
\label{subsec::datasets}

We curated a diverse training set comprising 10K samples from each of five publicly available training sets: ChartQA~\cite{ChartQA}, DocVQA~\cite{DocVQA}, TextVQA~\cite{TextVQA}, LLaVA-Multi~\cite{jiang2024mantis}, and VisionThink-Smart~\cite{VisionThink}. We also collected 2k samples with the same distribution as a validation set. The SFT phase using a subset of the training set (5k samples from each dataset), while the GRPO uses all collected examples. This mixture spans both document understanding and natural image domains.

We evaluate on six benchmarks spanning natural image understanding-RealWorldQA~\cite{RealWorldQA} (real-world spatial reasoning), POPE~\cite{POPE} (object hallucination), and $V^{*}$-Bench~\cite{Vstar} (visual search in high-resolution scenes)-and document understanding-ChartQA~\cite{ChartQA} (reasoning over charts), DocVQA~\cite{DocVQA} (forms, tables, letters, and handwriting), and OCRBench~\cite{OCRBench} (text recognition and text-centric reasoning). full details are in the appendix.

\subsection{Implementation Details}
\label{subsec::impl_details}
We conduct experiments based on Qwen2.5-VL-7B-Instruct~\cite{qwen2_5_vl}; all compared methods are built on the same base VLM to isolate the impact of efficiency mechanisms.

Unless otherwise specified, each sample is first processed at a low-resolution setting $I_{\mathrm{low}}$ has hight and width devided by 2, corresponding to the ``LR'' baseline in Table~\ref{tab:main_results} (RTR=0.25).
When a crop is requested, the crop image(s) are rendered at the native high-resolution token density of the base model.

We use the discrete crop set $\mathcal{C}$ from \S\ref{sec:data_curation} (four quadrants, four half-image regions, a center crop, and the full image).
Requested crops are appended to the context together with $I_{\mathrm{low}}$.

For training, we use HuggingFace TRL~\cite{trl} for SFT and GRPO, adapting TRL's GRPO trainer to support multi-image and multi-turn conversations.
For the SFT phase we use a total batch size of 16 with learning rate $1\times 10^{-4}$ and LoRA rank 8. $w_t=5$ only for the tool-call turn and $1$ otherwise.
For the GRPO phase we use a total batch size of 64 with $G{=}8$ generations per sample, learning rate $1\times 10^{-5}$, and LoRA rank 8. $w_a = 10$, $\alpha_{miss}=2$, $\alpha_{\text{use}}=0.25$ and $\lambda=0.01$

\begin{table*}[t]
\centering
\caption{\textbf{Main results across vision-language benchmarks.} We compare \method{} against fixed-ratio token-pruning methods (VisionZIP, SparseVLM, Holo-V) and dynamic, crop-inspired methods (ERGO, DeepEyes, PixelReasoner, VisionThink). Retain Token Ratios (RTR) denote the fraction of visual tokens retained. ``LR'' indicates the first turn uses a low-resolution image. \method{} matches the full-resolution baseline (Qwen2.5-VL-7B) while using only 36\% of the visual tokens, outperforming all efficient alternatives. Best results per column in \textbf{bold}.}
\label{tab:main_results}
\resizebox{0.92\textwidth}{!}{%
\begin{tabular}{l| cc cc cc cc cc cc | cc}
\toprule
& \multicolumn{2}{c}{ChartQA} & \multicolumn{2}{c}{DocVQA} & \multicolumn{2}{c}{OCRBench} & \multicolumn{2}{c}{POPE} & \multicolumn{2}{c}{RealWorld} & \multicolumn{2}{c}{V$^{*}$ Bench} & \multicolumn{2}{c}{\textbf{Average}} \\
\cmidrule(lr){2-3} \cmidrule(lr){4-5} \cmidrule(lr){6-7} \cmidrule(lr){8-9} \cmidrule(lr){10-11} \cmidrule(lr){12-13} \cmidrule(lr){14-15}
Model & Acc $\uparrow$ & RTR $\downarrow$ & Acc $\uparrow$ & RTR $\downarrow$ & Acc $\uparrow$ & RTR $\downarrow$ & Acc $\uparrow$ & RTR $\downarrow$ & Acc $\uparrow$ & RTR $\downarrow$ & Acc $\uparrow$ & RTR $\downarrow$ & Acc $\uparrow$ & RTR $\downarrow$ \\
\midrule
\multicolumn{15}{l}{\textit{Base models}} \\
Qwen2.5-VL-7B    & 79.80 & 1.00 & 94.00 & 1.00 & 81.10 & 1.00 & 87.87 & 1.00 & 68.80 & 1.00 & 71.20 & 1.00 & 80.46 & 1.00 \\
Qwen2.5-VL-7B-LR & 65.00 & 0.25 & 91.00 & 0.25 & 70.70 & 0.25 & 84.41 & 0.25 & 66.00 & 0.25 & 63.20 & 0.25 & 73.39 & 0.25 \\
\midrule
\multicolumn{15}{l}{\textit{Token-pruning methods (fixed ratio)}} \\
Holo-V (50\%)    & 62.04 & 0.50 & 70.77 & 0.50 & 68.20 & 0.50 & 86.54 & 0.50 & 67.19 & 0.50 & 64.40 & 0.50 & 69.86 & 0.50 \\
Holo-V (70\%)    & 69.32 & 0.70 & 76.40 & 0.70 & 72.40 & 0.70 & 87.37 & 0.70 & \textbf{68.89} & 0.70 & 69.11 & 0.70 & 73.92 & 0.70 \\
SparseVLM (50\%) & 73.20 & 0.50 & 83.60 & 0.50 & 75.60 & 0.50 & 85.50 & 0.50 & 68.40 & 0.50 & 54.45 & 0.50 & 73.46 & 0.50 \\
SparseVLM (70\%) & 75.80 & 0.70 & 87.20 & 0.70 & 79.30 & 0.70 & 85.40 & 0.70 & 68.50 & 0.70 & 54.45 & 0.70 & 75.11 & 0.70 \\
VisionZIP (50\%) & 74.76 & 0.50 & 89.39 & 0.50 & 69.20 & 0.50 & 87.56 & 0.50 & 66.01 & 0.50 & 66.49 & 0.50 & 75.57 & 0.50 \\
VisionZIP (70\%) & 76.72 & 0.70 & 90.75 & 0.70 & 72.70 & 0.70 & 87.86 & 0.70 & 64.84 & 0.70 & 65.97 & 0.70 & 76.47 & 0.70 \\
\midrule
\multicolumn{15}{l}{\textit{Dynamic / crop-inspired methods}} \\
ERGO         & 80.30 & 1.293 & 59.80 & 0.435 & 74.60 & 1.597 & 86.20 & 1.277 & 65.50 & 0.933 & 83.80 & 0.401 & 75.03 & 0.99 \\
ERGO (LR)    & 73.40 & 0.614 & 60.40 & 0.380 & 72.10 & 0.840 & 84.40 & 0.577 & 64.30 & 0.466 & 81.20 & 0.366 & 72.63 & 0.54 \\
DeepEyes     & 61.76 & 1.090 & 82.95 & 1.012 & 80.50 & 1.350 & 34.40 & 1.260 & 65.75 & 1.090 & 80.10 & 1.001 & 67.58 & 1.13 \\
DeepEyes (LR)& 51.04 & 0.277 & 82.04 & 0.253 & 66.80 & 0.529 & 32.74 & 0.296 & 61.05 & 0.277 & 72.25 & 0.253 & 60.99 & 0.31 \\
PixelReasoner     & 87.90 & 1.010 & 92.93 & 1.020 & 70.80 & 1.020 & \textbf{88.30} & 1.130 & 54.20 & 1.000 & \textbf{84.81} & 1.050 & 79.82 & 1.04 \\
PixelReasoner (LR)& 74.00 & 0.270 & 92.15 & 0.270 & 59.30 & 0.300 & 85.06 & 0.400 & 51.76 & 0.250 & 70.15 & 0.270 & 72.07 & \textbf{0.29} \\
VisionThink  & 79.90 & 1.150 & 90.35 & 0.320 & 80.10 & 0.830 & 86.70 & 0.340 & 66.60 & 0.550 & 71.73 & 0.490 & 79.23 & 0.61 \\
\midrule
\method{}    & \textbf{80.64} & 0.320 & \textbf{94.43} & 0.280 & \textbf{81.30} & 0.420 & 85.73 & 0.270 & 68.50 & 0.430 & 71.20 & 0.420 & \textbf{80.30} & 0.36 \\
\bottomrule
\end{tabular}
}
\end{table*}

\subsection{Baselines}
\label{subsec::baselines}
All baselines are based on Qwen-2.5-VL-7B~\cite{qwen2_5_vl}. \textbf{Fixed-budget token pruning/compression:} VisionZip~\cite{VisionZip}, SparseVLM~\cite{SparseVLM}, and Holo-V~\cite{HoloV}, training-free inference-time methods configured to retain $50\%$ or $70\%$ of the full-resolution visual tokens. \textbf{Dynamic methods:} VisionThink~\cite{VisionThink}, which escalates a low-resolution pass to the \emph{full} high-resolution image, and the reasoning-based ERGO, Deep-Eyes, and Pixel-Reasoner~\cite{lee2025ergo, DeepEyes, wang2025pixel}, which can request a crop from it. \method{} instead escalates \emph{spatially}, via a short tool call. 

\subsection{Main Results}
\label{subsec::main_results}

Table~\ref{tab:main_results} reports task performance and efficiency.
\method{} achieves an average score of 80.30, matching the full-resolution baseline (80.46) while using only $0.36\times$ the visual tokens on average.
This indicates that the learned \emph{coupled-decision policy} answers directly from the low-resolution overview when possible, and invokes targeted crops only when necessary.

Compared to fixed-budget pruning methods, \method{} consistently attains higher accuracy at comparable or lower RTR.
For example, at $70\%$ retained tokens, VisionZip trails \method{} by over 4 percent on average (76.47 vs.\ 80.30), reflecting the limitation of a global, sample-agnostic token budget.

Against the adaptive escalation baseline VisionThink~\cite{VisionThink}, \method{} improves both accuracy (80.30 vs.\ 79.23) and efficiency (RTR 0.36 vs.\ 0.61).
Notably, on ChartQA \method{} slightly exceeds the full-resolution baseline (+0.84 percent) while reducing RTR to 0.32, whereas VisionThink often performs two visual passes and exceeds baseline compute (RTR=1.15).

\subsubsection{Beyond Token Efficiency}
Although RTR captures the dominant visual compute, end-to-end latency also depends on the number of autoregressive text tokens generated: other dynamic methods decide whether to escalate through extended reasoning traces, which inflates decoding cost even when the final answer is short, whereas \method{} encodes the decision in a short, structured tool call and reuses the KV cache between turns.

\begin{figure}
  \centering
  \includegraphics[width=0.95\linewidth]{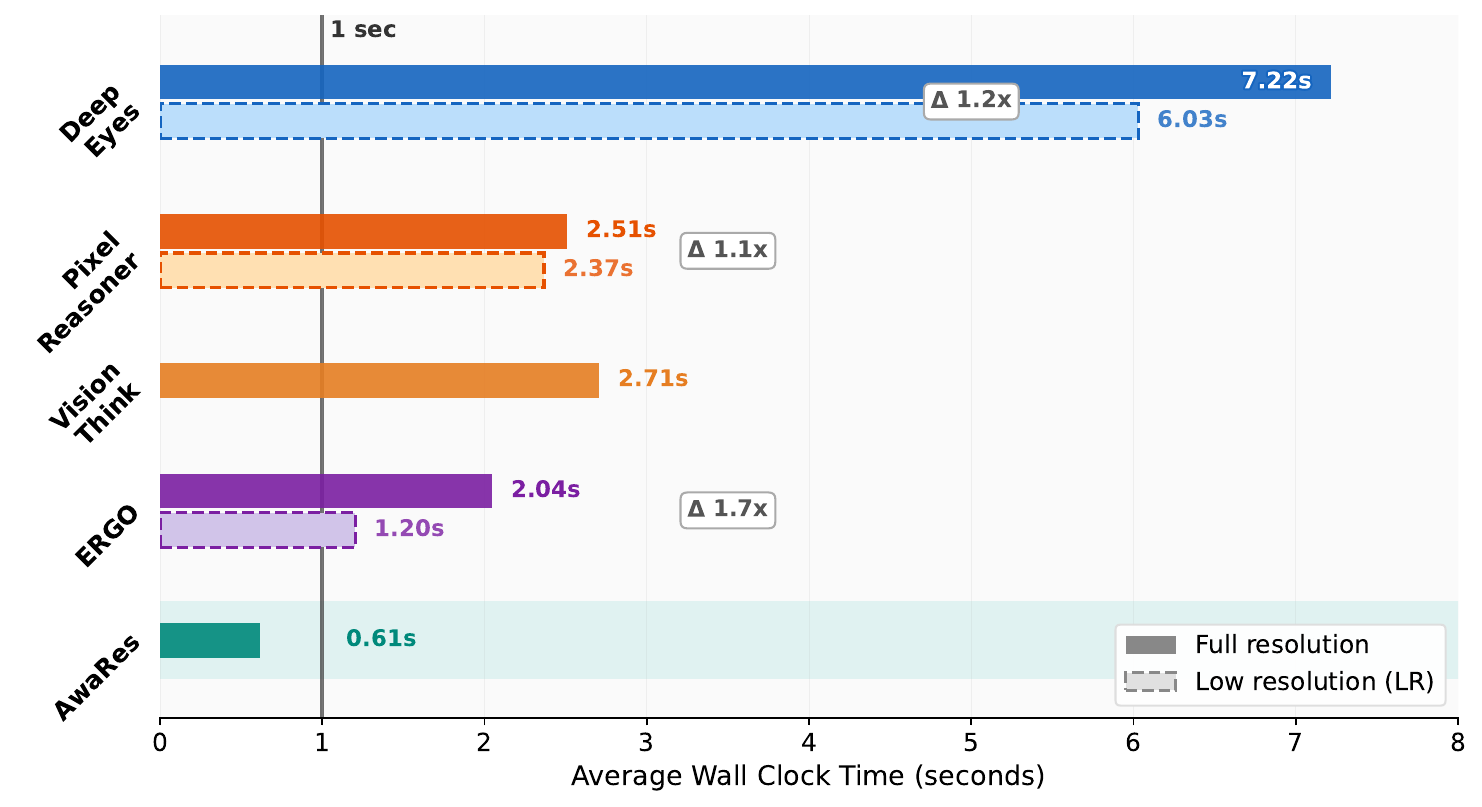}
  \caption{\textbf{Latency breakdown.} Full vs. low-resolution (LR) first-turn input; solid bars are full-resolution wall-clock time, dashed bars the LR variant. The small speedups ($\Delta 1.1$--$1.7\times$) show that input token count is not the main bottleneck: cost is dominated by extended reasoning or iterative cropping. AwaRes reaches 0.61s via a single short tool call, avoiding verbose generation.}
  \label{fig:performance_wallclock}
  \end{figure}

Fig.~\ref{fig:performance_wallclock} reports average wall-clock time for all dynamic methods, with a reference line at one second-a practical requirement for interactive applications. \method{} is the only method consistently below it, averaging 0.61s across the six benchmarks, and is $2\textrm{--}12\times$ faster than every baseline. The figure also reports each baseline's self-speedup from a low-resolution first turn (``LR''): ERGO benefits most ($1.7\times$, 2.04\,s\,$\rightarrow$\,1.20\,s), still nearly twice \method{}'s latency, while PixelReasoner and DeepEyes gain only $1.1\times$ and $1.2\times$, indicating their latency is dominated by text decoding and multi-turn overhead rather than visual encoding. Two choices drive this: the resolution decision costs a single decoding step rather than a reasoning trace, and the loop terminates early when the low-resolution view already suffices, avoiding a second forward pass. Under a production-grade engine (vLLM), \method{} averages 0.24s per sample, 31\% faster than the high-resolution baseline, so the token savings do translate into deployment latency.


\begin{figure}
    \includegraphics[width=0.85\linewidth]{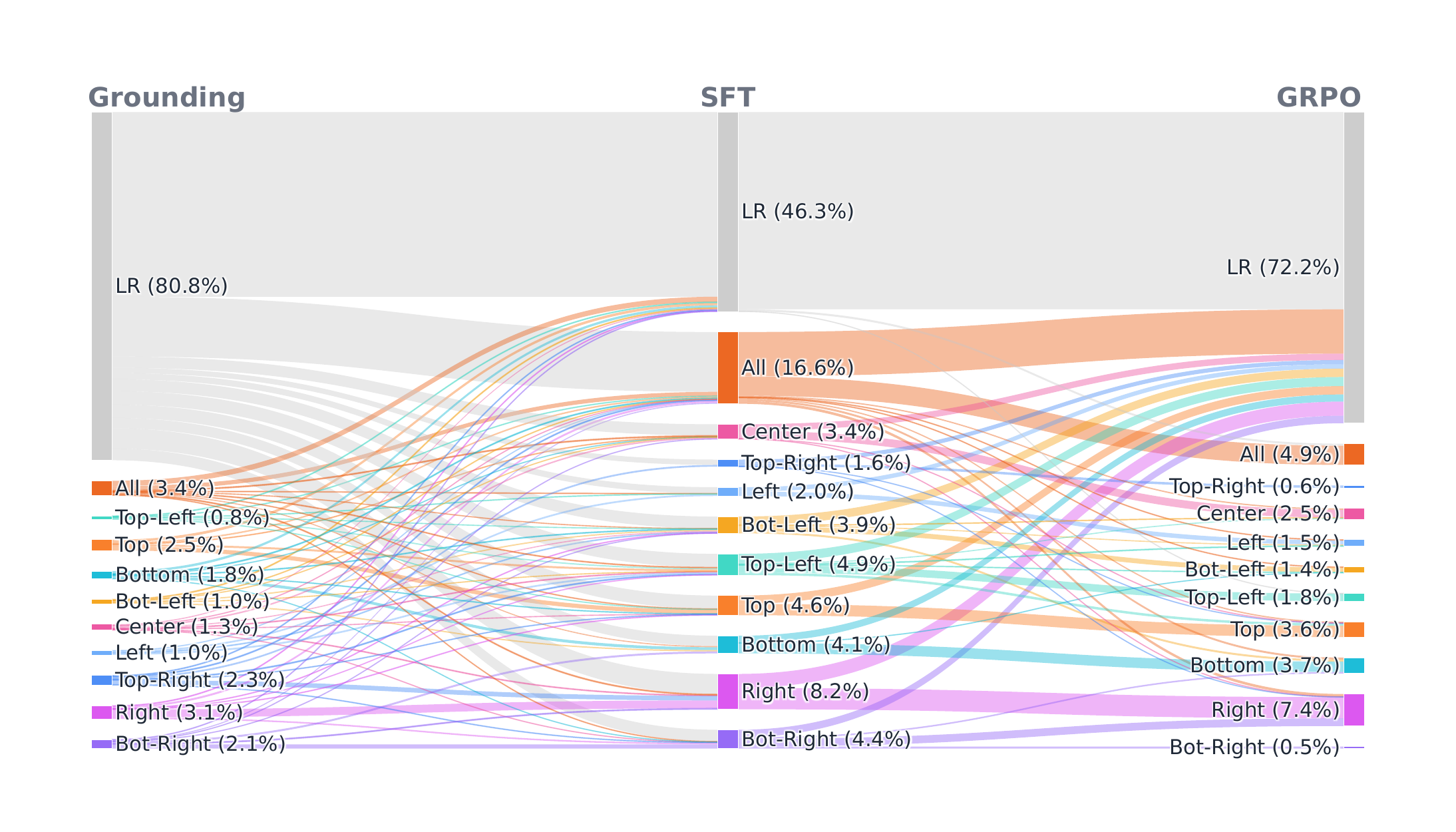}
    \caption{\textbf{From Over-Using to Looking Where It Matters.} The flow of crop selection from Grounding (Oracle GT for the SFT), SFT-tuned model (middle), and GRPO predictions (right). SFT, designed to introduce the tool-calling protocol, tends to over-use the crop tool. The GRPO phase corrects this behavior through the tool-use cost, increasing low-resolution decisions while exploring alternative crop strategies that balance accuracy and efficiency.}
    \label{fig:crop_flow_between_phases}
\end{figure}

\subsubsection{From Over-Calling to Looking Where It Matters}
Fig.~\ref{fig:crop_flow_between_phases} illustrates how the model's \emph{first-turn crop-request policy} evolves from grounded supervision (denoted as Oracle GT) through SFT to GRPO.
The SFT designed to introduce the tool-calling protocol, and as so, the policy becomes conservative with respect to accuracy and therefore \emph{over-invokes}{} the crop tool: the probability of taking the no-call action (LR) drops sharply (80.8\%$\rightarrow$46.3\%), while escalation to the full-image crop (``All'') is substantially inflated (3.4\%$\rightarrow$16.6\%).
This behavior is consistent with imitation-style training that prioritizes adhering to the tool protocol, even in cases where the low-resolution view is sufficient.

GRPO then reshapes the same policy under the explicit accuracy--efficiency objective in Eq.~\ref{eq:reward}; as a result, the policy shifts toward selective tool use: LR increases to 72.2\% and ``All'' decreases to 4.9\%, closer to the grounding model.
Due to the tool-use cost that penalize crop size, the learned policy may shift towards efficient strategies that differ from the grounded annotations.

We further show that \method{} does not merely distill the grounding model: the 7B model surpasses the tool-prompted grounding model (MoE, 235B-A22B) on ChartQA ($80.6$ vs. $69.7$) and OCRBench ($81.3$ vs. $77.8$), indicating that GRPO discovers cropping strategies beyond those introduced by the grounding model (full results in the Appendix).

\subsection{Ablations}
\label{subsec::ablations}
We ablate the data-preparation pipeline, the cold-start SFT stage, and the GRPO stage. The full details are in Appendix~\ref{sec:supp_sft} and~\ref{sec:grpo_ablations}.

\subsubsection{Data preparation}
\label{sec:data_preparation_ablation}
Swapping the default judge \textsc{LLaMA-3.3-70B}~\cite{llama3} for \textsc{DeepSeek-V3.2}~\cite{liu2025deepseek} changes only $3.12\%$ of resolution-selection labels, versus $19.89\%$ for ANLS, which over-penalizes paraphrased-but-correct answers and costs $3.8$ points when used to train $\pi_{\theta_{\mathrm{ref}}}$. Two independent grounders agree on $88\%$ of cases.

\subsubsection{Tool introduction}
Table~\ref{tab:supp_sft_ablation_full} studies how SFT recipes affect the emergence of reliable crop-tool usage. The default optimizes each assistant message independently, whereas \textbf{Traj} performs trajectory-level SFT over the entire two-turn interaction as a single sample, and \textbf{Phased} trains on the tool-call turns first. Trajectory-level training improves both accuracy and efficiency over split training (77.90 vs.\ 75.15 and 0.43 vs.\ 0.36), and raising the tool-turn weight to $w_t{=}5$ improves accuracy further (79.70) at a higher RTR (0.49), suggesting that emphasizing the first-turn decision encourages more confident tool invocation. \textbf{Upsampled} variants reduce RTR (to 0.33) but also accuracy, reflecting the expected trade-off. We therefore initialize GRPO from trajectory-level SFT with $w_t{=}5$.

\subsubsection{Tool Optimization rewards}
We ablate the GRPO reward components, which combine a sentence-transformer answer-similarity reward with a tool-use cost (Table~\ref{tab:supp_grpo_ablation}). Removing the crop-area cost raises RTR from 0.36 to 0.42 ($\lambda=0$), and removing the crop cost entirely raises it to 0.51; the increase is bounded by the SFT initialization and the KL term, which keep the policy close to a reference that already uses the tool sensibly. ANLS and LLM-as-a-judge correctness rewards perform similarly, since we optimize short answers rather than lengthy reasoning traces; sentence-transformer similarity balances the semantic sensitivity ANLS lacks against the robustness an LLM judge lacks.

\subsubsection{The effect of cold start initialization}
An alternative to our two-stage approach is to run RL directly from the base model and let it learn tool use from scratch during GRPO. Trained this way for 3 epochs on the same data (GRPO only, Tab.~\ref{tab:supp_grpo_ablation}), the model barely uses the tool, falls back to low-resolution processing, and loses accuracy. SFT alone reaches good accuracy but high RTR from over-calling, and it is a double-edged sword: on an external validation set (Fig.~\ref{fig:crop_flow_between_phases}) the SFT model has overfitted to the training data and shifted its policy substantially. GRPO is therefore not only exploratory but also recovers from that overfitting.

\section{Conclusion}
\label{sec:conclusions}
We presented \method{}, a spatial-on-demand inference framework for VLMs that keeps a low-resolution global view and retrieves high-resolution crops only when the query requires them, through a deployment-friendly tool-calling interface with multi-turn KV-cache reuse. Training needs no manual spatial supervision: an automatic curation pipeline supplies the data, and we train with cold-start SFT followed by multi-turn GRPO under a reward combining answer correctness and crop-usage penalties. Across six benchmarks, \method{} matches full-resolution accuracy with far fewer visual tokens and lower latency than both dynamic and fixed-budget baselines.

\section*{Limitations}
Our framework acquires high-resolution evidence from a \emph{fixed, discrete}
crop set (four quadrants, four half-image regions, a center crop, and the full image). This keeps the tool interface structured and deployment-friendly, but it
also caps spatial precision: when the answer-bearing evidence straddles crop
boundaries or occupies a region much smaller than a quadrant, the policy must
request more area than strictly necessary. Extending the interface to continuous
bounding boxes is left to future work.

 The resolution-sufficiency label comes from an LLM judge and
the crop target from a grounding model; both make mistakes
(Appendix~\ref{sec:supp_data_curation} shows representative failures), and the
resulting label noise propagates into the cold-start policy. We measure
inter-model agreement rather than validating against human annotations.

We evaluate a single VLM (Qwen2.5-VL-7B) and latency numbers are measured on a single
NVIDIA H100-80GB GPU under HuggingFace and vLLM, meaning absolute wall-clock times
may differ on other serving stacks.

\clearpage
\bibliography{main}

\appendix
\input{appendix}

\end{document}

%% file: preamble.tex
\newcommand{\yes}{\textbf{v}}
\newcommand{\no}{\textbf{x}}

\newcommand{\method}{AwaRes}

%% file: appendix.tex

This supplementary document provides additional details, analyses, and visual examples that complement the main paper. We organize the material as follows: Section~\ref{sec:supp_data_curation} presents supplementary visual examples from our data curation pipeline, including both successful annotations and failure cases of the automatic process. In addition, we analyze alternative data curation strategies using ANLS-based filtering, along with grounding model robustness. Section~\ref{sec:benchmark_examples} provides qualitative examples of \method{} at inference time, illustrating both successful predictions and failure modes. Section~\ref{sec:latency_details} details our latency analysis, including hardware setup, wall-clock measurements comparing \method{} and VisionThink, and a response length analysis explaining the observed efficiency gains. Section~\ref{sec:supp_sft} offers an in-depth analysis of the cold-start (SFT) stage, examining the coupled-decision policy (CDP) behavior, sensitivity to the tool-turn weight $w_t$, and tool-call formatting reliability. Section~\ref{sec:grpo_ablations} presents GRPO ablations covering the cold-start stage, tool-cost penalty, and correctness reward, as well as how GRPO learns beyond grounding model annotations. Section~\ref{sec:training_details} provides comprehensive training hyperparameters and configuration details. Finally, Section~\ref{sec:prompts} includes the complete prompts used throughout our pipeline: the LLM-as-a-Judge prompt for data curation, the grounding model localization prompt for grounding, and the system prompt used during SFT, GRPO, and inference.

\section{Data Annotation}
\label{sec:supp_data_curation}
\subsection{Data Curation Pipeline}
\begin{figure*}[t]
\centering
\begin{tabular}{c}
\includegraphics[width=0.9\textwidth]{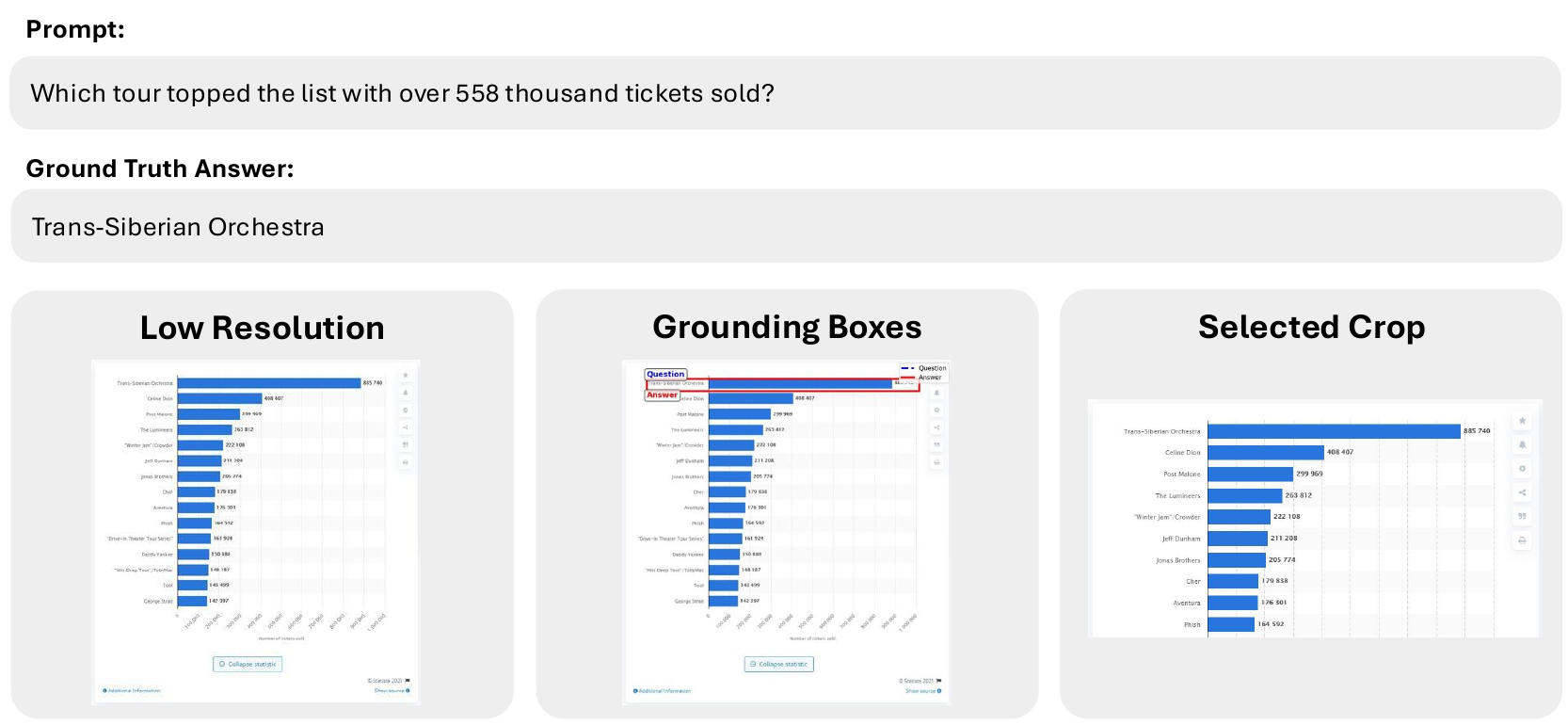} \\
\midrule
\includegraphics[width=0.9\textwidth]{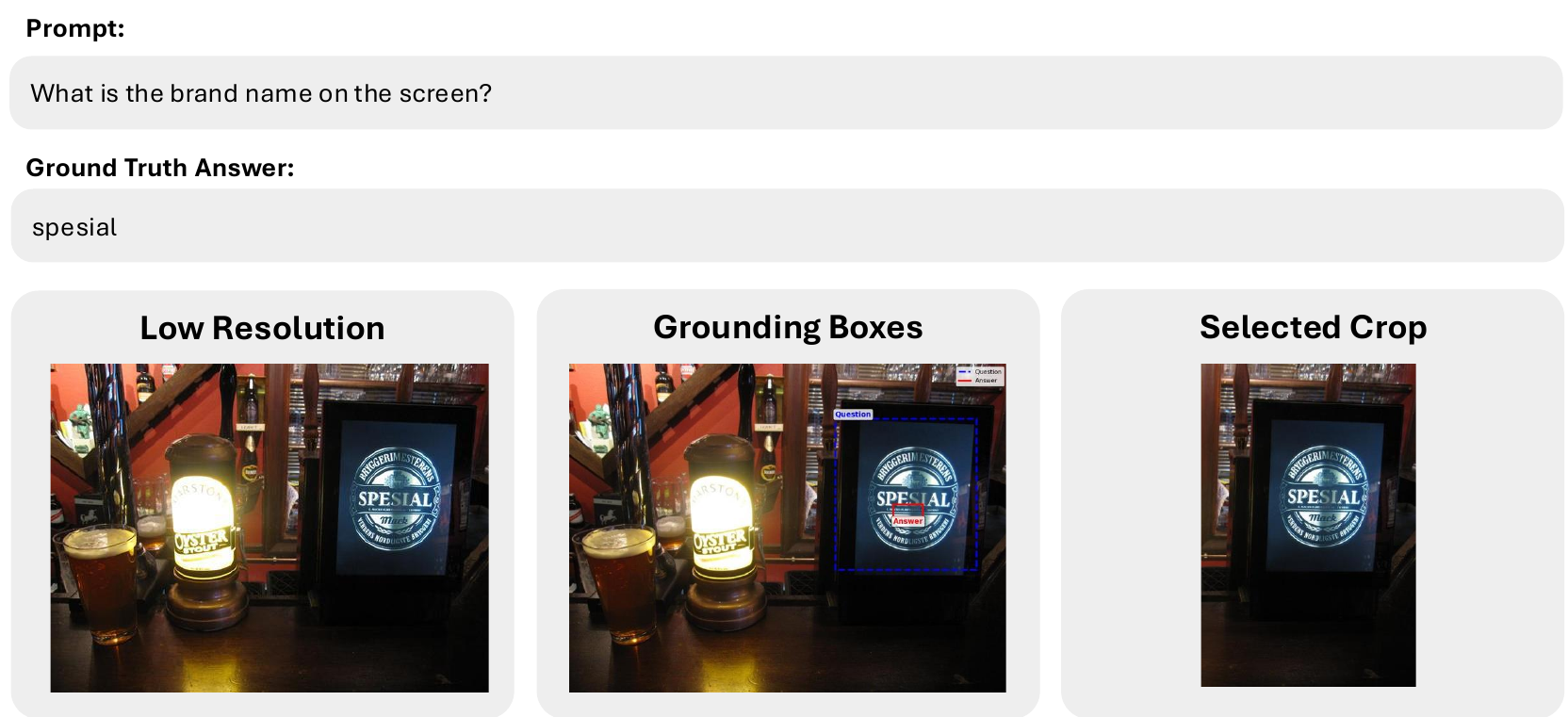} \\
\end{tabular}
\caption{Visual examples from our data curation pipeline. Each row shows the low-resolution input image, the grounding model-detected bounding boxes (question region in blue, answer region in red), and the selected crop used for training. Top: A Chart example where the grounding model correctly localizes the relevant bar and its label. Bottom: A natural image example requiring OCR of a brand banner.}
\label{fig:data_curation_examples_2}
\end{figure*}

\begin{figure*}[t]
\centering
\begin{tabular}{c}
\includegraphics[width=0.9\textwidth]{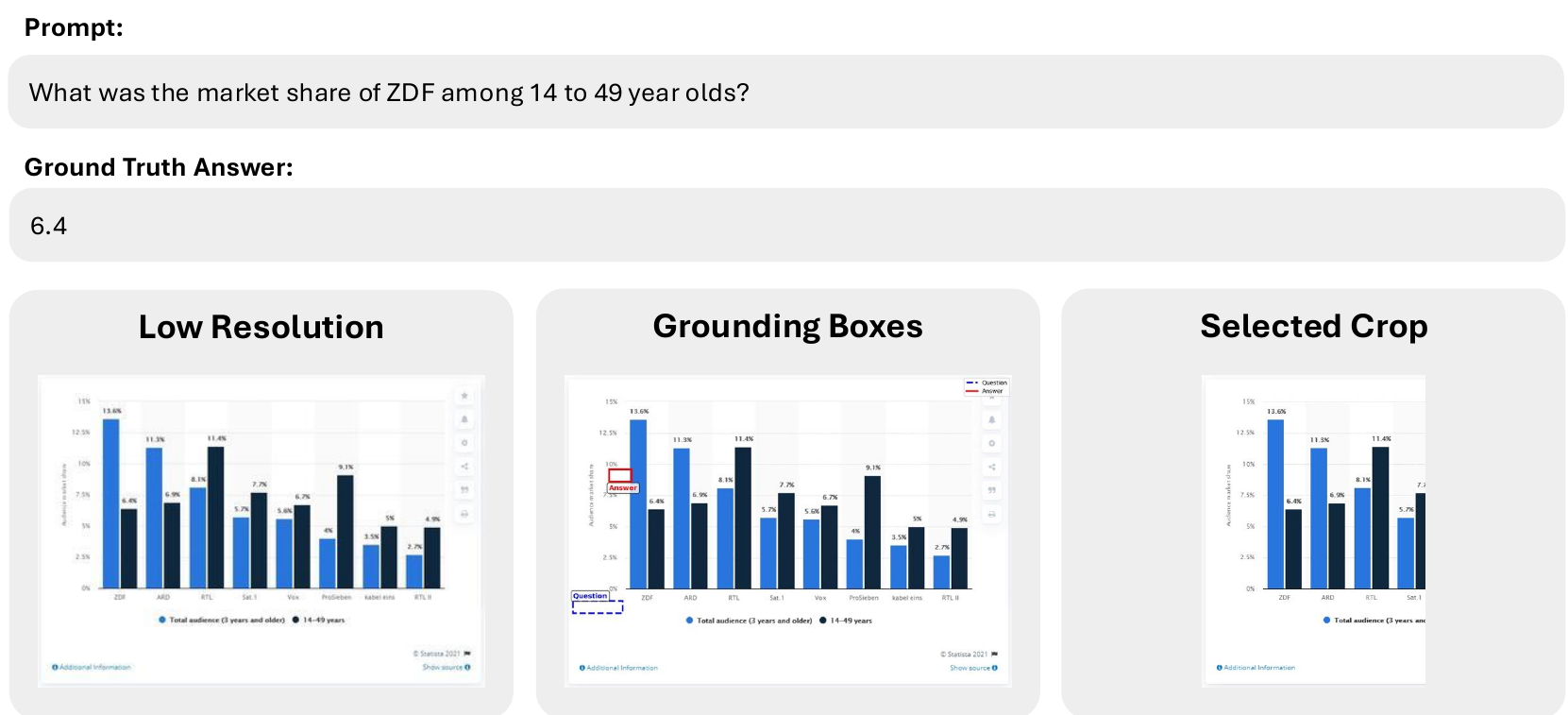} \\
\midrule
\includegraphics[width=0.9\textwidth]{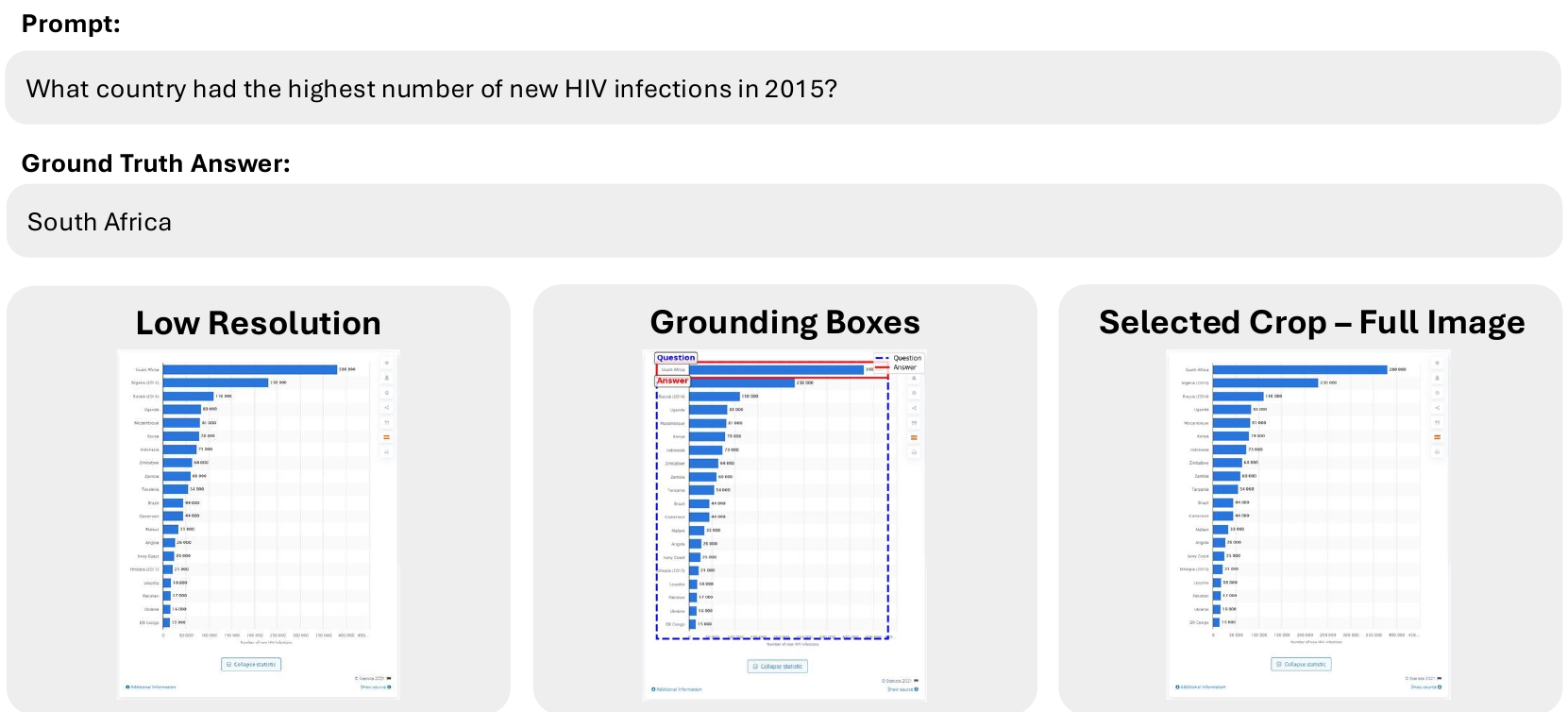} \\
\end{tabular}
\caption{Failure cases from the automatic data curation pipeline. \textbf{Top:} The grounding model incorrectly localizes the answer region, missing the relevant bar (ZDF's 14-49 age group value). However, the crop regions helps mitigate such localization errors-the selected crop still contains the correct answer. \textbf{Bottom:} The grounding model correctly identifies the question and answer regions but the bounding boxes span nearly the entire image, resulting in a full-image crop selection. While this produces correct supervision, it is inefficient as no resolution savings are achieved.}
\label{fig:data_curation_examples_bad}
\end{figure*}

In this section, we provide in Figure~\ref{fig:data_curation_examples_2} supplementary visual examples from our data curation pipeline. Additionally, in Figure~\ref{fig:data_curation_examples_bad} we illustrate failure cases where the automatic process did not produce satisfactory results.

\subsection{Agreement of resolution selection}
The two panels compare resolution-selection labels from our default judge \textsc{LLaMA-3.3-70B}~\cite{llama3} against \textsc{DeepSeek-V3.2}~\cite{liu2025deepseek} (left) and ANLS (right), with LR and HR marking low- and high-resolution labels. The two LLM judges show strong agreement ($96.88\%$ along the diagonal), whereas ANLS agrees far less ($80.11\%$). This indicates that many responses are semantically equivalent despite differing in phrasing, which ANLS penalizes - strengthening our decision to use a judge rather than a simple text-comparison metric.
  \begin{figure}
  \centering
  \includegraphics[width=0.6\linewidth]{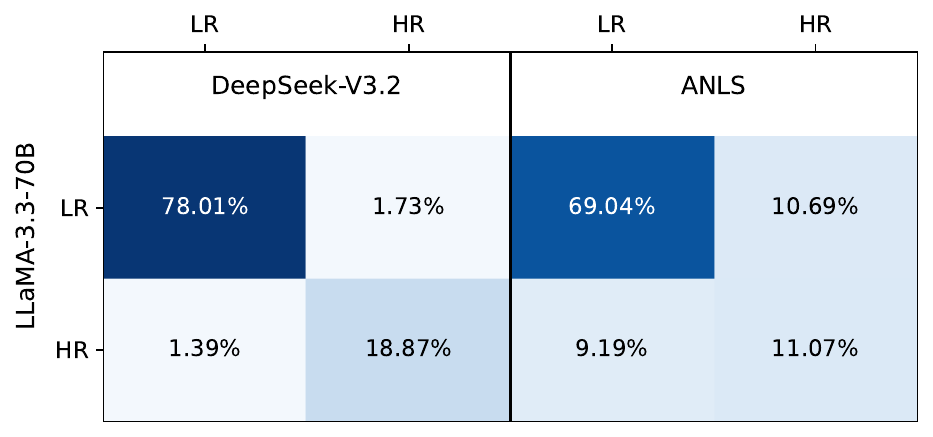}
  \caption{\textbf{Agreement on resolution-selection labels.}}
  \label{fig:judge_robustness}
  \end{figure}

\subsection{Training data statistics}
\begin{figure}[t]
\centering
\includegraphics[width=\linewidth]{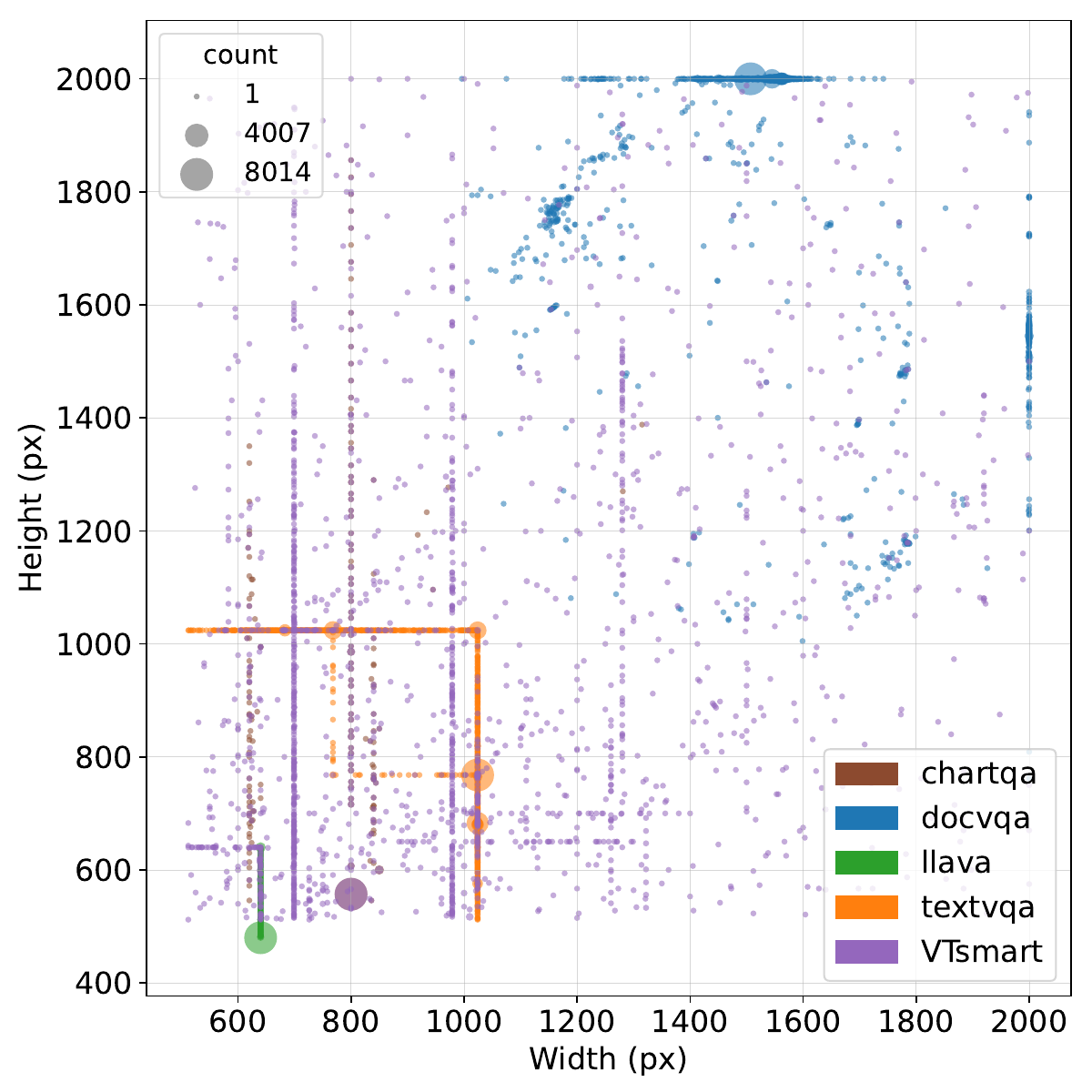}
\caption{Resolution distribution of training datasets. Each point represents an image's width and height in pixels.}
\label{fig:trainig_res_stats}
\end{figure}

Figure~\ref{fig:trainig_res_stats} shows the resolution distribution of our training data. DocVQA~\cite{DocVQA} exhibits the highest resolutions, whereas the LLaVA-Multi~\cite{jiang2024mantis} subset contains the lowest. VisionThink-Smart~\cite{VisionThink} and ChartQA~\cite{ChartQA} display a broad spread across resolutions, while TextVQA concentrations cluster around 1000 pixels along one axis. We cap all resolutions at 2000$\times$2000, as prior work~\cite{CARES} demonstrated that truncating DocVQA resolution has negligible impact on performance.

\subsection{ANLS-Based Data Curation Analysis}
\label{subsec::anls_analysis}
We compare annotating the crops using LaaJ vs.\ using ANLS in Tab.~\ref{tab:label_strategy_comparison}.
ANLS is a string-oriented similarity measure designed for OCR-style exactness, and as such it is ill-suited for supervising our resolution-sufficiency labels, specifically since we observe the VLM answers with crops act as augmented input and thus perturbed the answers a bit (even when they are semantically correct).

This effect is reflected empirically.
Training with ANLS-based labels for cold-start yields substantially lower average accuracy than LaaJ-based labels (75.9 vs.\ 79.70), with pronounced drops on text- and detail-sensitive benchmarks such as ChartQA (70.5 vs.\ 77.0) and OCRBench (70.0 vs.\ 78.8).
While ANLS labeling attains a lower RTR (0.38 vs.\ 0.49), the accompanying accuracy degradation indicates that it often suppresses crop requests even when additional resolution is required, rather than producing a better-calibrated policy.
Overall, LaaJ provides a more reliable semantic correctness signal for annotation, which is critical for learning the CDP without over-penalizing benign surface-form variation.

\begin{table*}[t]
\centering
\small
\renewcommand{\arraystretch}{1.15}
\caption{Comparing labeling strategies: LaaJ (LLaMA-3.3-70B) vs.\ ANLS. We report \textbf{Accuracy $\uparrow$} and \textbf{RTR $\downarrow$} across all benchmarks.}
\label{tab:label_strategy_comparison}
\resizebox{\textwidth}{!}{%
\begin{tabular}{l | cc cc cc cc cc cc | cc}
\toprule
\textbf{Label Strategy} & \multicolumn{2}{c}{ChartQA} & \multicolumn{2}{c}{DocVQA} & \multicolumn{2}{c}{OCRBench} & \multicolumn{2}{c}{POPE} & \multicolumn{2}{c}{RealWorld} & \multicolumn{2}{c}{V$^{*}$ Bench} & \multicolumn{2}{c}{\textbf{Average}} \\
\cmidrule(lr){2-3} \cmidrule(lr){4-5} \cmidrule(lr){6-7} \cmidrule(lr){8-9} \cmidrule(lr){10-11} \cmidrule(lr){12-13} \cmidrule(lr){14-15}
 & Acc $\uparrow$ & RTR $\downarrow$ & Acc $\uparrow$ & RTR $\downarrow$ & Acc $\uparrow$ & RTR $\downarrow$ & Acc $\uparrow$ & RTR $\downarrow$ & Acc $\uparrow$ & RTR $\downarrow$ & Acc $\uparrow$ & RTR $\downarrow$ & Acc $\uparrow$ & RTR $\downarrow$ \\
\midrule
ANLS                  & 70.5 & 0.31 & 93.6 & 0.28 & 70.0 &0.41  & 86.7 &0.27 & 69.0 &0.47  & 65.9 & 0.54 & 75.9 & 0.38 \\
LLaMA-3.3-70B (LaaJ)  & \textbf{77.0} & 0.42 & \textbf{94.0} & 0.35 & \textbf{78.8} & 0.61 & \textbf{88.0} & 0.32 & 69.7 & 0.64 & \textbf{70.7} & 0.60 & \textbf{79.70} & 0.49 \\
\bottomrule
\end{tabular}}
\end{table*}

\subsection{Grounding model robustness}
\label{subsec::oracle_robustness}
\begin{figure}
\centering
\includegraphics[width=\linewidth]{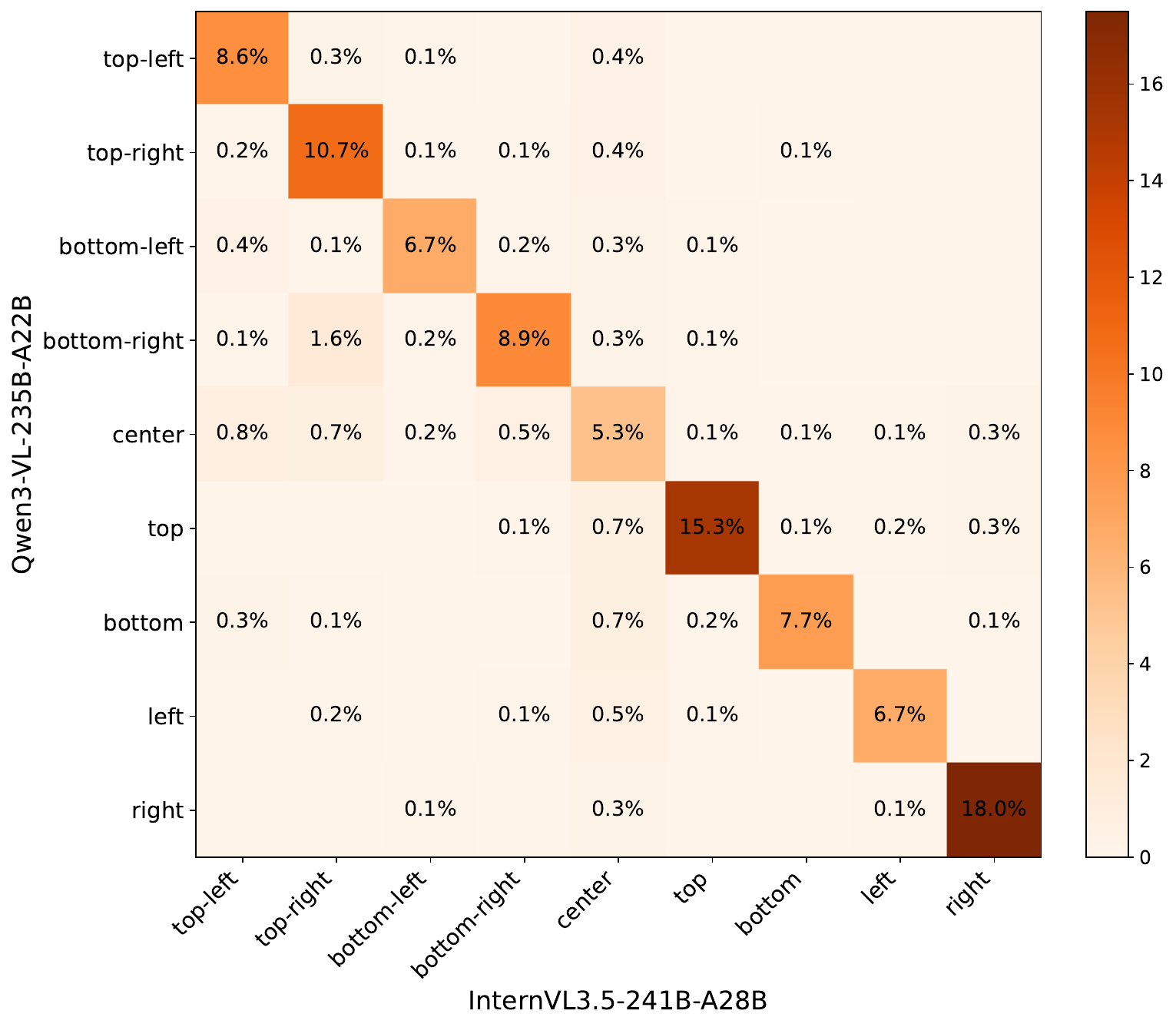}
\caption{Crop confusion matrix between the two grounding model, conditioned on samples where high-resolution cropping helps. The diagonal concentration shows 88\% agreement between the grounding models on the selected crop region.}
\label{fig:crop-confusion-matrix}
\end{figure}

To assess the reliability of our grounding model, we compare the crop regions selected by two strong VLM grounding modelS, Qwen3-VL-235B-A22B~\cite{qwen3_vl} and InternVL3.5-241B-A28B~\cite{wang2025internvl3}, on the subset of samples where high-resolution cropping is beneficial. As shown in Figure~\ref{fig:crop-confusion-matrix}, the two grounding models exhibit strong agreement, with 88\% of predictions concentrated along the diagonal, indicating that the selected crop regions are largely model-agnostic rather than artifacts of a single grounding model's idiosyncrasies. The few off-diagonal cases cluster among spatially adjacent regions (e.g., bottom-right vs.\ top-right), suggesting that disagreements stem from boundary ambiguity rather than fundamentally divergent localization.

\section{Benchmark Qualitative Examples}
\label{sec:benchmark_examples}
\begin{figure*}[t]
    \centering
    \setlength{\tabcolsep}{1pt}
    \renewcommand{\arraystretch}{0.6}
    \begin{tabular}{c}
        \includegraphics[width=0.72\textwidth]{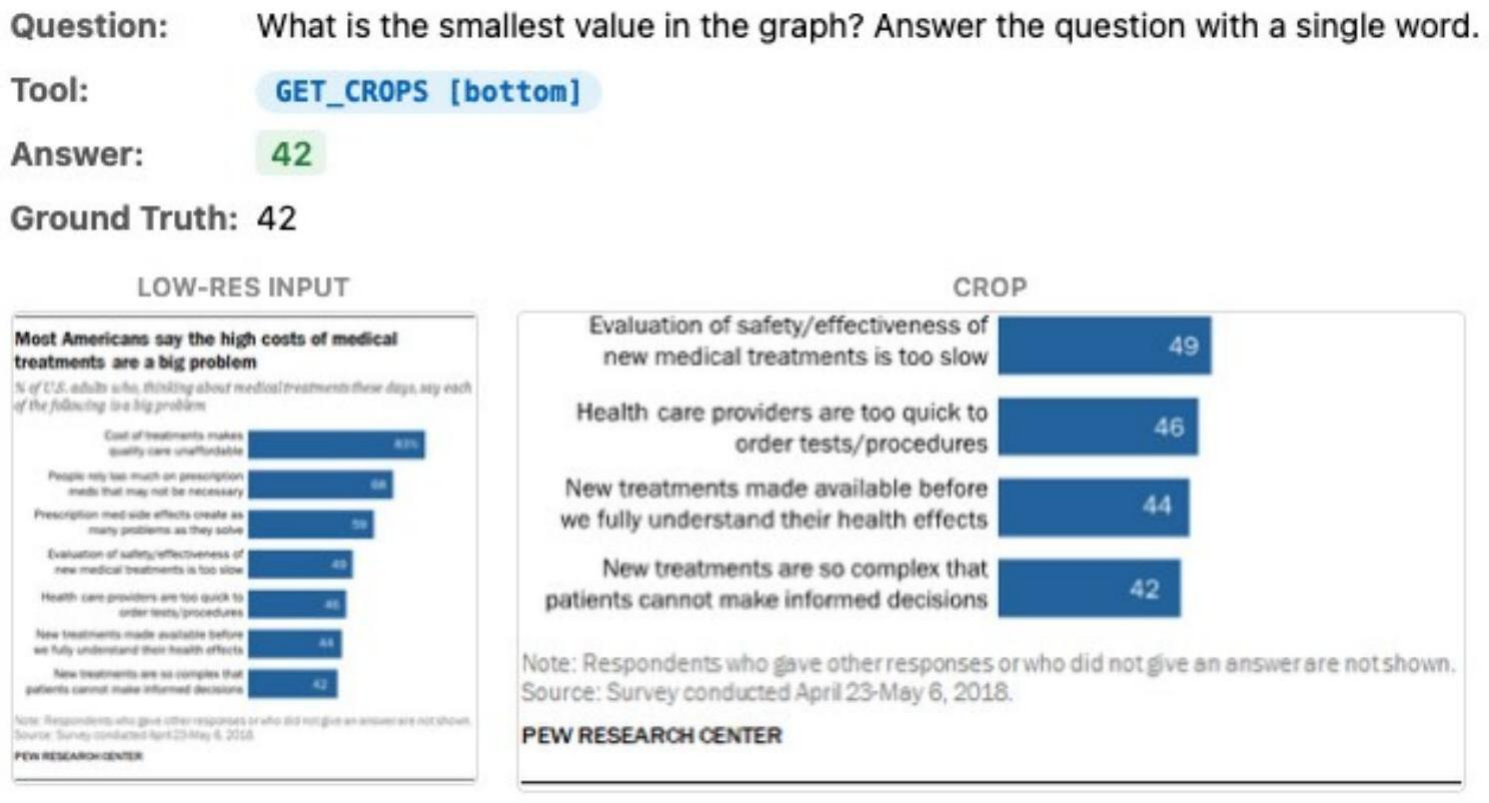} \\[1pt]
        \midrule \\[-4pt]
        \includegraphics[width=0.72\textwidth]{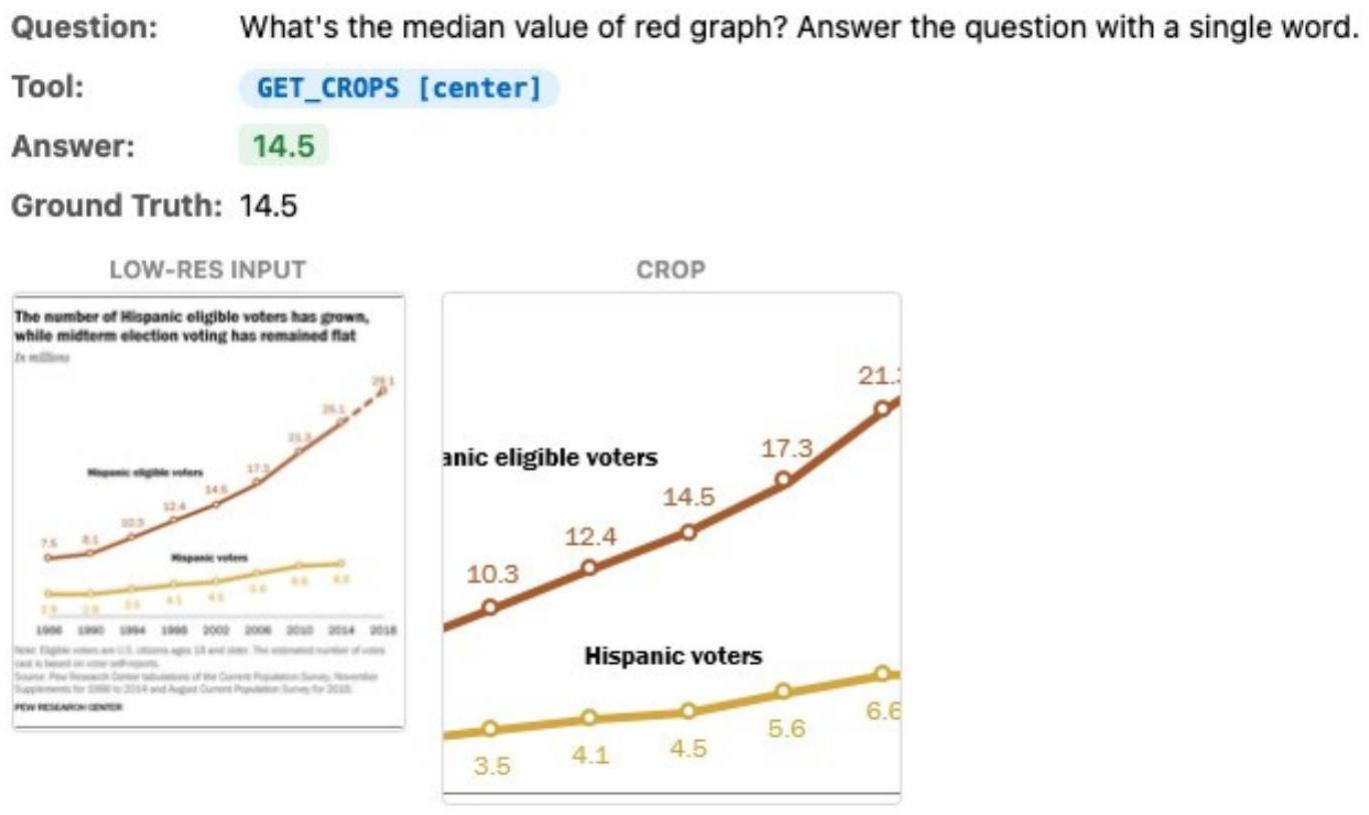} \\[1pt]
        \midrule \\[-4pt]
        \includegraphics[width=0.72\textwidth]{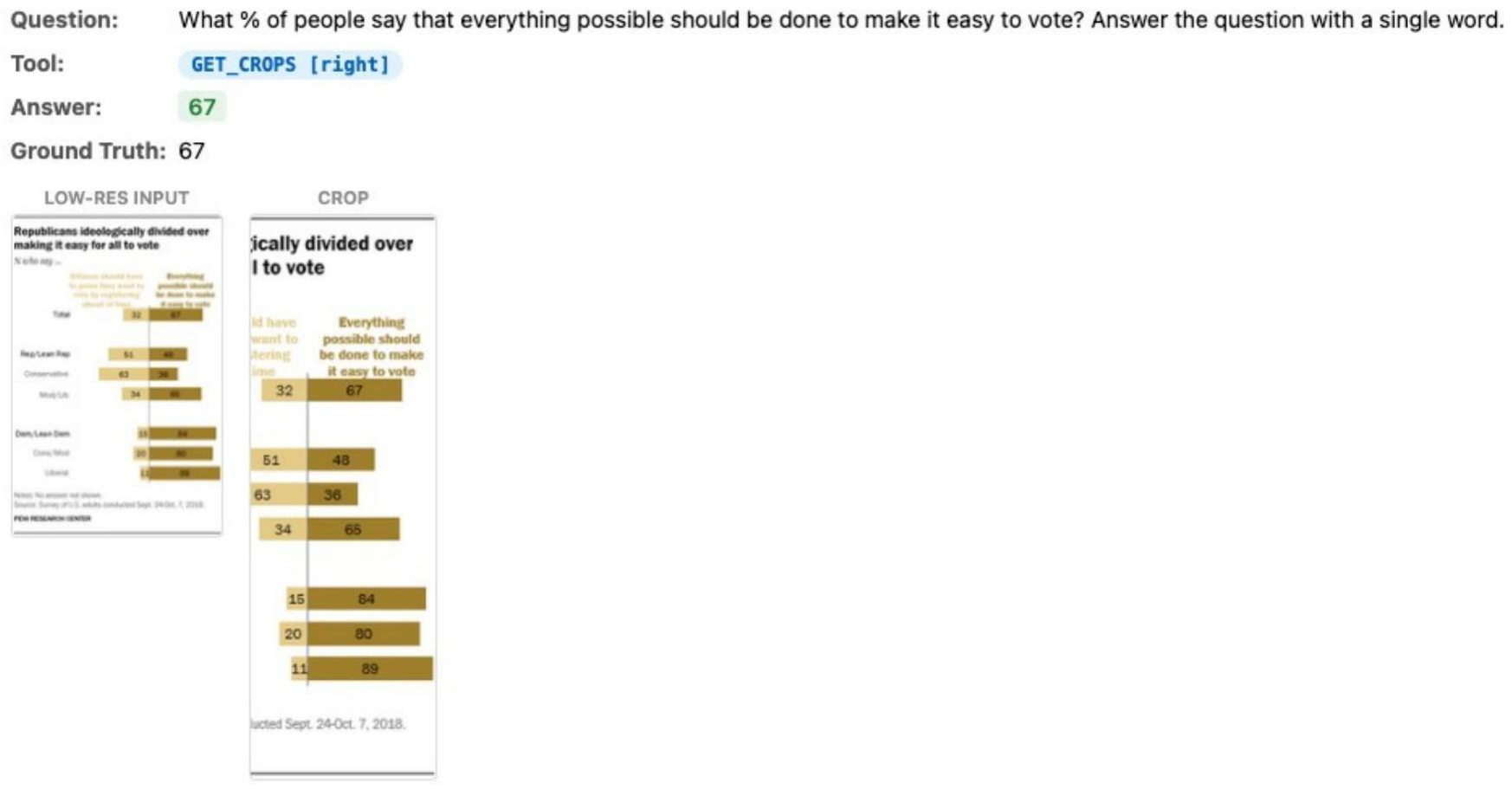} \\
    \end{tabular}
    \caption{\textbf{Positive examples of \method{}'s adaptive cropping (1/6).}  The model isolates the relevant bar segments and axis labels to identify the smallest chart value, zooms into data points along a trend line to compute a median, and focuses on labeled chart categories to read a specific percentage, extracting precise numerical values that are indiscernible in the low-resolution overview alone.}
    \label{fig:positive_examples_1}
\end{figure*}

\begin{figure*}[t]
    \centering
    \setlength{\tabcolsep}{1pt}
    \renewcommand{\arraystretch}{0.6}
    \begin{tabular}{c}
        \includegraphics[width=0.9\textwidth]{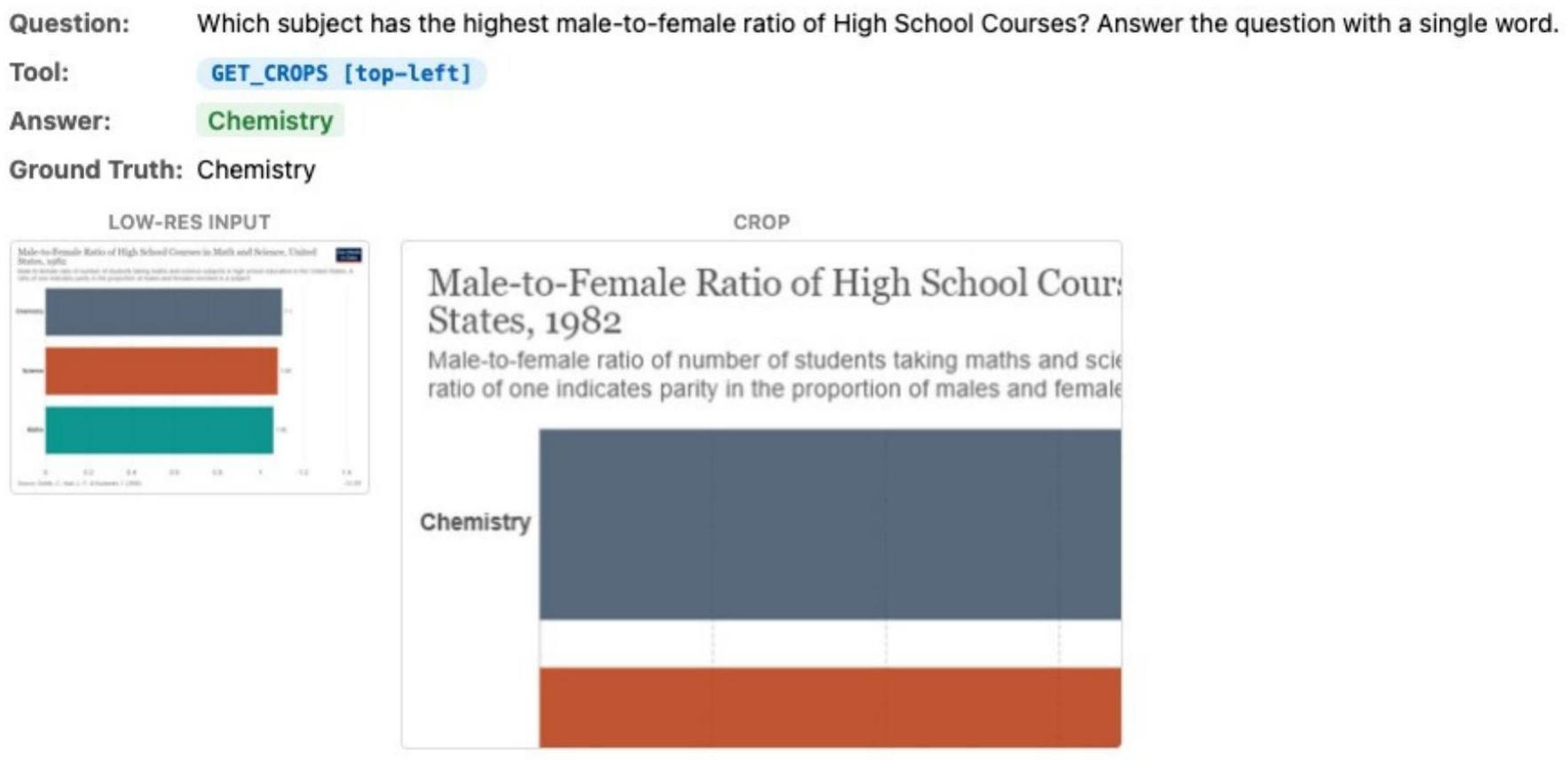} \\[1pt]
        \midrule \\[-4pt]
        \includegraphics[width=0.9\textwidth]{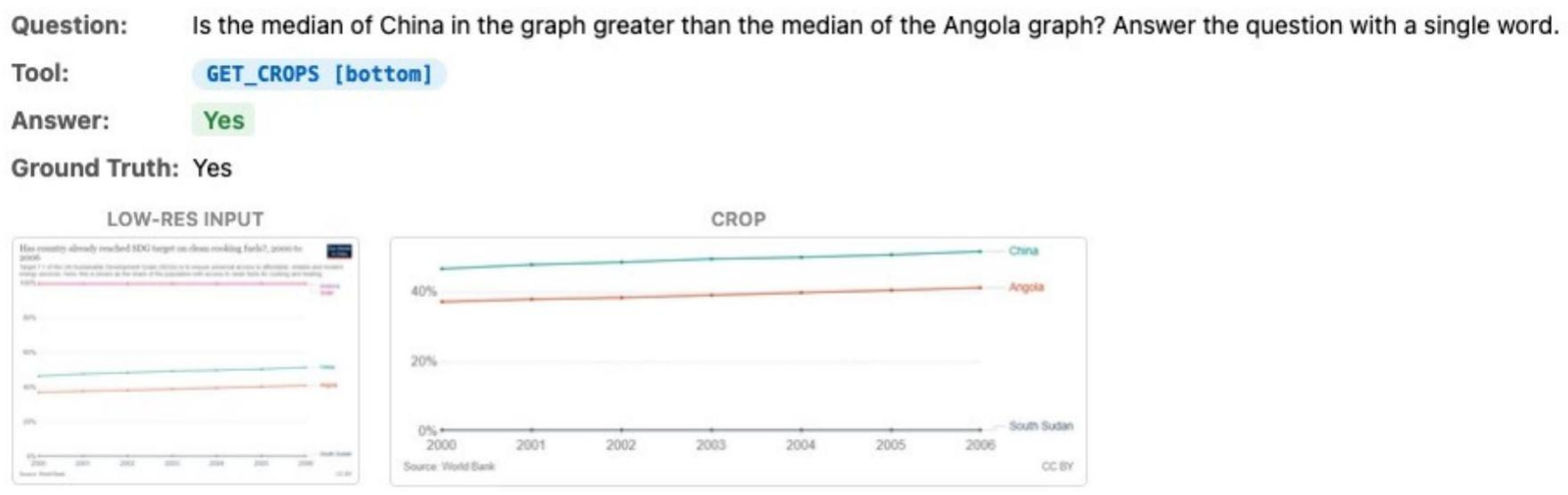} \\[1pt]
        \midrule \\[-4pt]
        \includegraphics[width=0.9\textwidth]{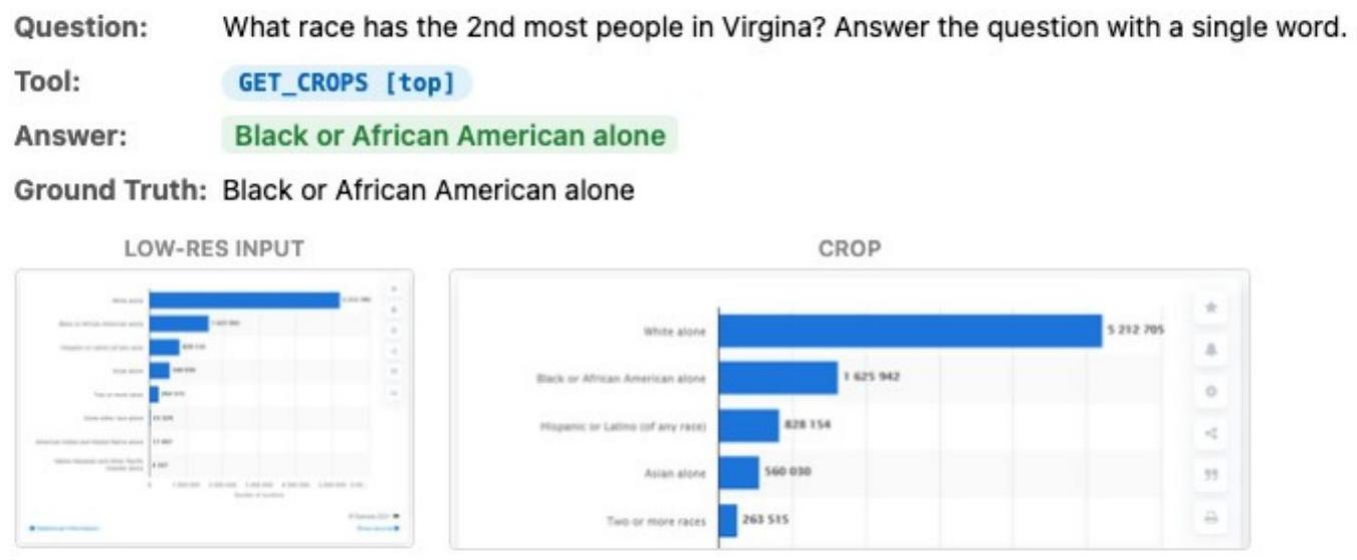} \\
    \end{tabular}
    \caption{\textbf{Positive examples of \method{}'s adaptive cropping (2/6).} The crops target the top bar and its label to identify the subject with the highest ratio, zoom into line-chart trends and country labels to compare medians across countries, and focus on the top entries of a horizontal bar chart to read demographic rankings. In each case, the high-resolution crop brings the answer-relevant region into sharp focus.}
    \label{fig:positive_examples_2}
\end{figure*}

\begin{figure*}[t]
    \centering
    \setlength{\tabcolsep}{1pt}
    \renewcommand{\arraystretch}{0.6}
    \begin{tabular}{c}
        \includegraphics[width=0.9\textwidth]{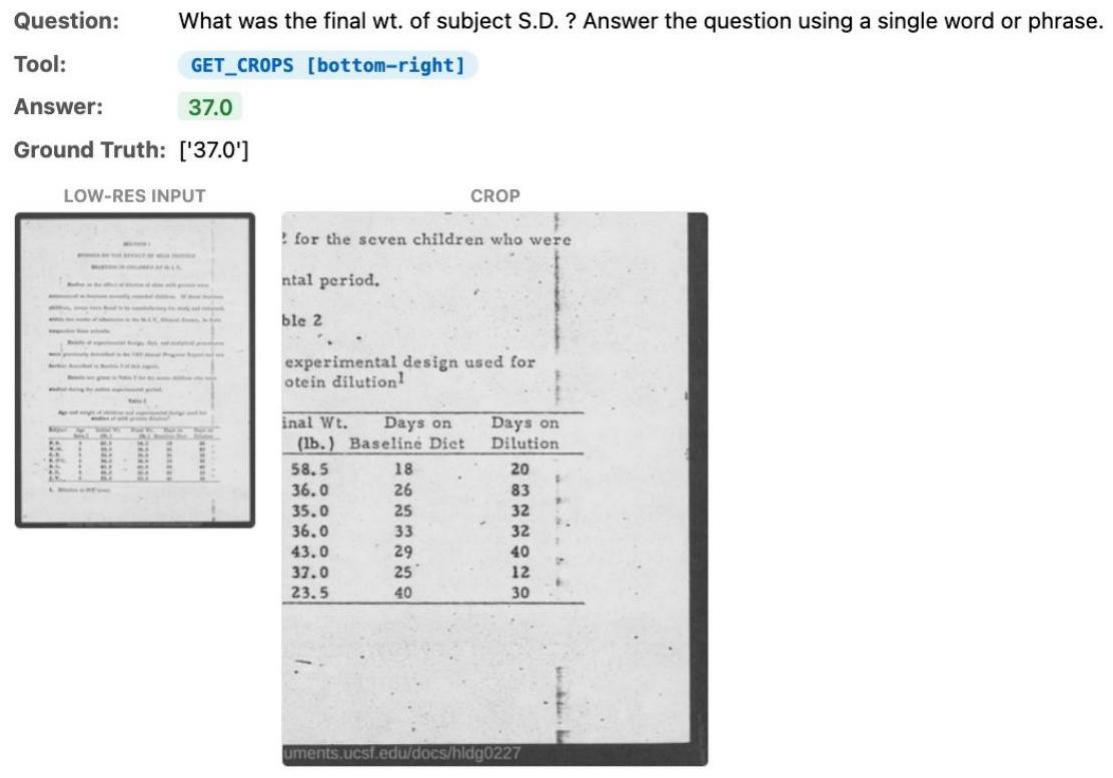} \\[1pt]
        \midrule \\[-4pt]
        \includegraphics[width=0.9\textwidth]{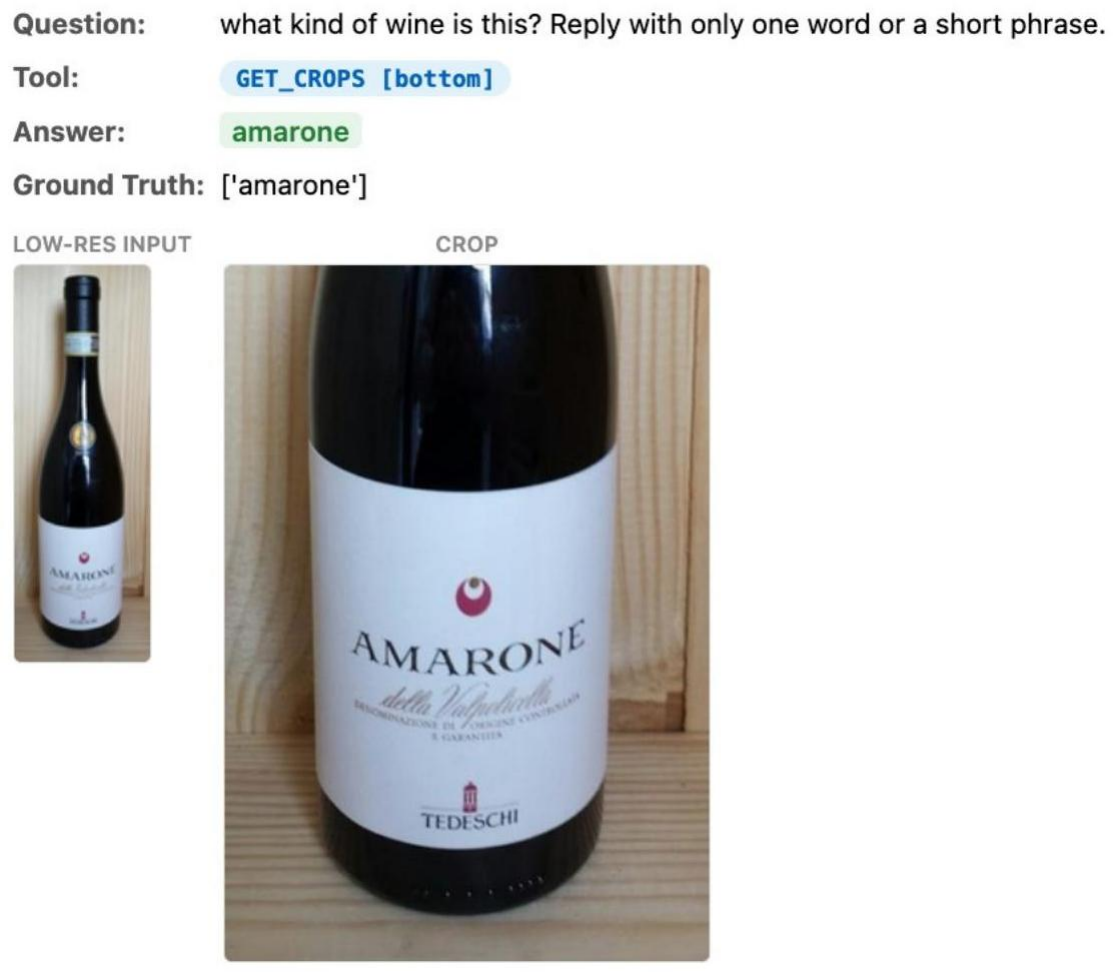} \\
    \end{tabular}
    \caption{\textbf{Positive examples of \method{}'s adaptive cropping (3/6).} The model crops into table cells of a scanned document to extract a specific weight value, and zooms into a wine bottle label to read the variety name. These examples demonstrate how targeted crops resolve fine-grained text on documents and labels that lack of OCR capabilities at low resolution.}
    \label{fig:positive_examples_3}
\end{figure*}

\begin{figure*}[t]
    \centering
    \setlength{\tabcolsep}{1pt}
    \renewcommand{\arraystretch}{0.6}
    \begin{tabular}{c}
        \includegraphics[width=0.9\textwidth]{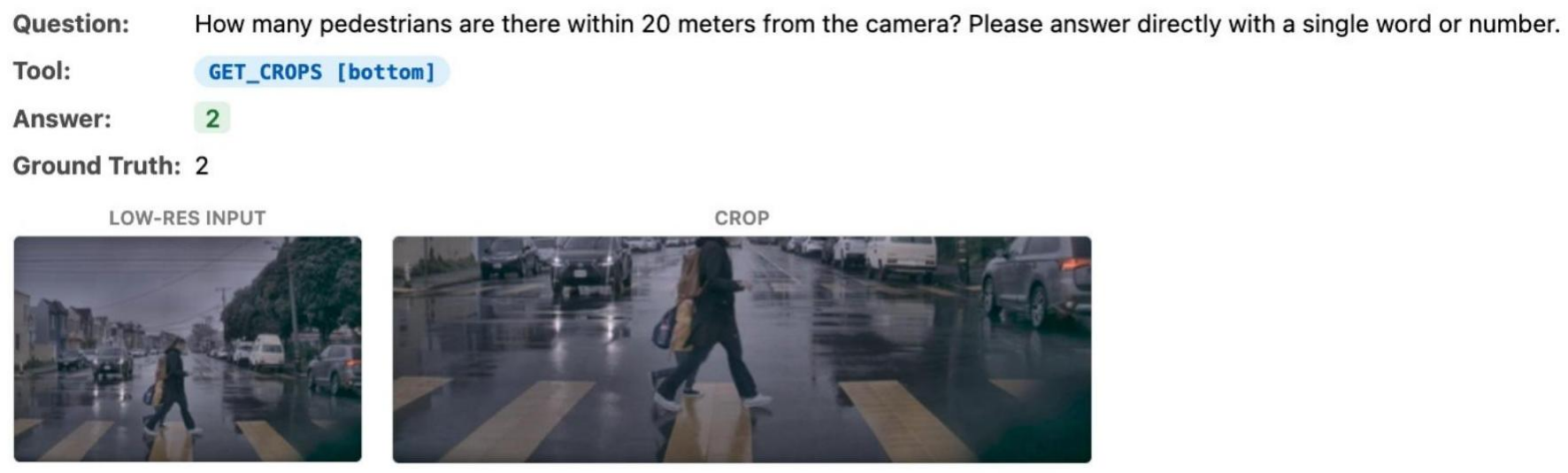} \\[1pt]
        \midrule \\[-4pt]
        \includegraphics[width=0.9\textwidth]{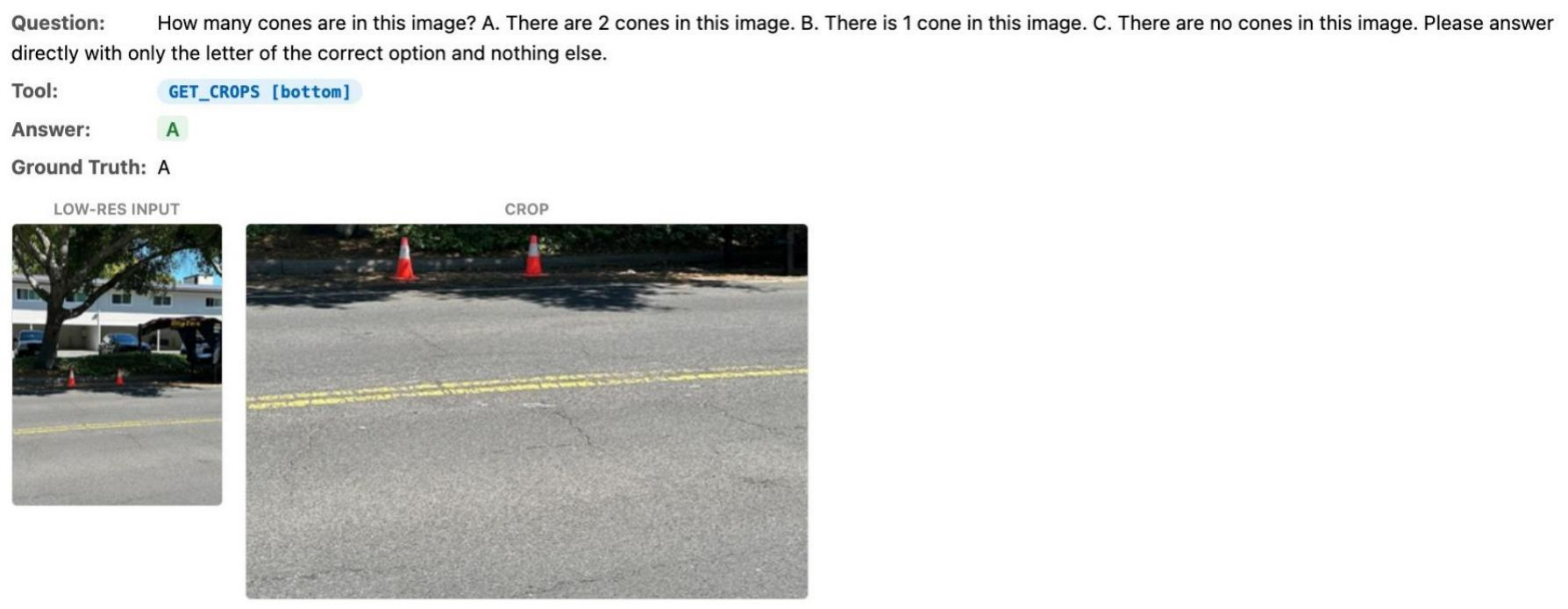} \\[1pt]
        \midrule \\[-4pt]
        \includegraphics[width=0.9\textwidth]{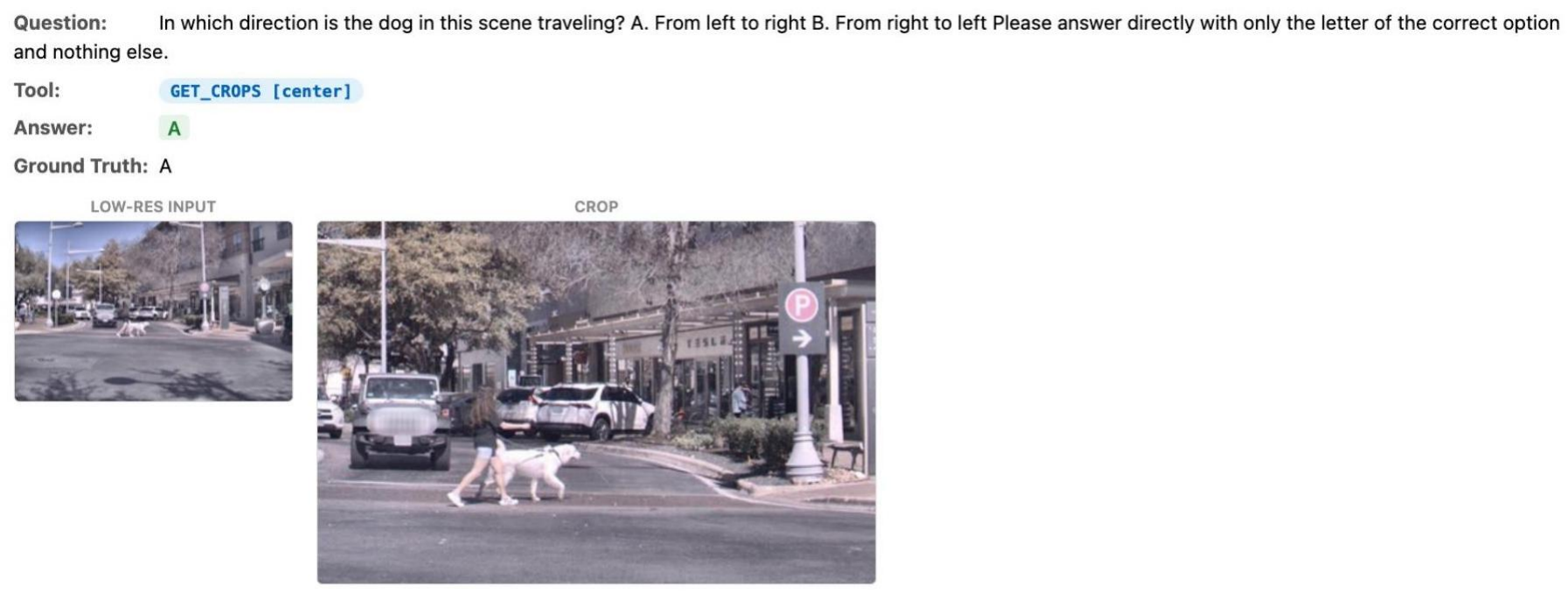} \\
    \end{tabular}
    \caption{\textbf{Positive examples of \method{}'s adaptive cropping (4/6).} The model correctly crops to count pedestrians in a rainy street scene, identify and count traffic cones in a parking lot, and determine the travel direction of a dog in an urban setting. The crops consistently center on the objects referenced by the question, enabling accurate spatial reasoning and counting from the high-resolution view of outdoor scenarios.}
    \label{fig:positive_examples_4}
\end{figure*}

\begin{figure*}[t]
    \centering
    \setlength{\tabcolsep}{1pt}
    \renewcommand{\arraystretch}{0.6}
    \begin{tabular}{c}
        \includegraphics[width=0.9\textwidth]{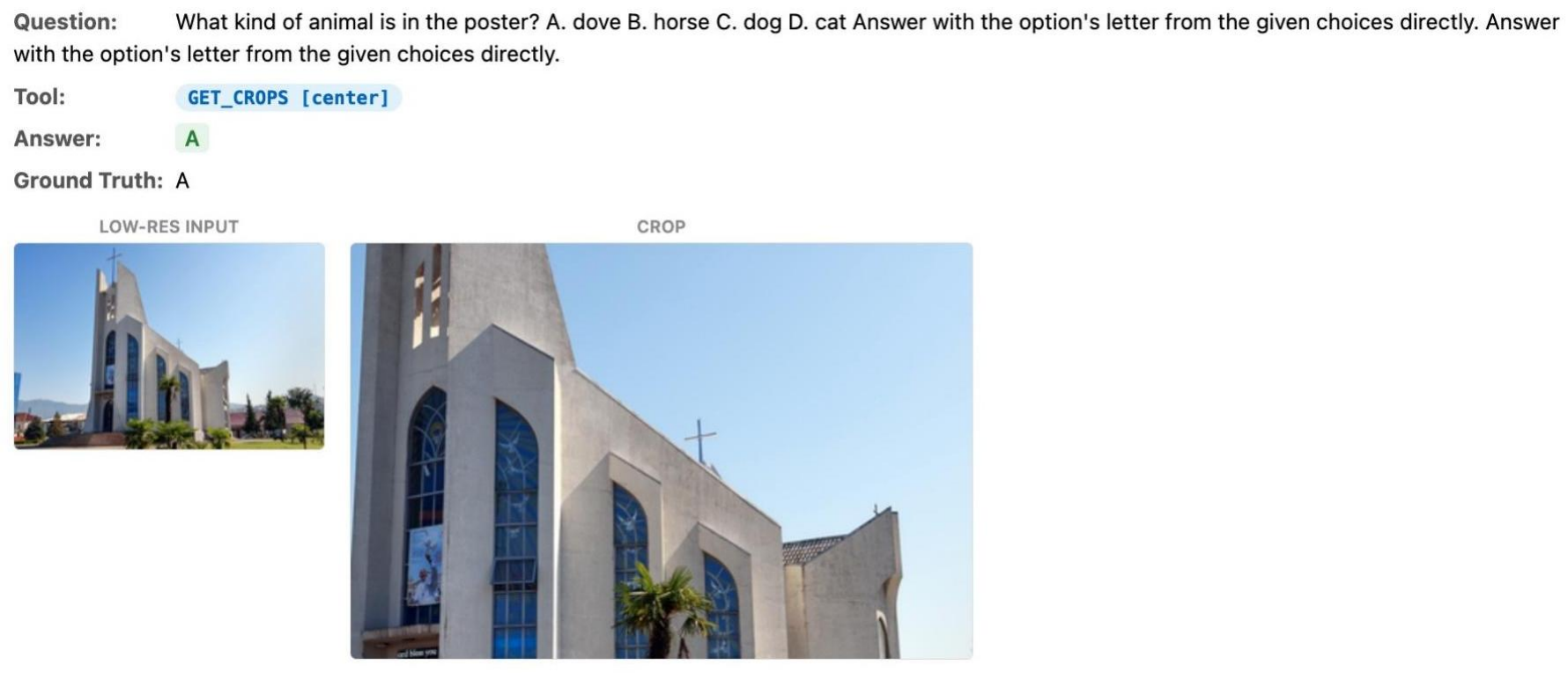} \\[1pt]
        \midrule \\[-4pt]
        \includegraphics[width=0.9\textwidth]{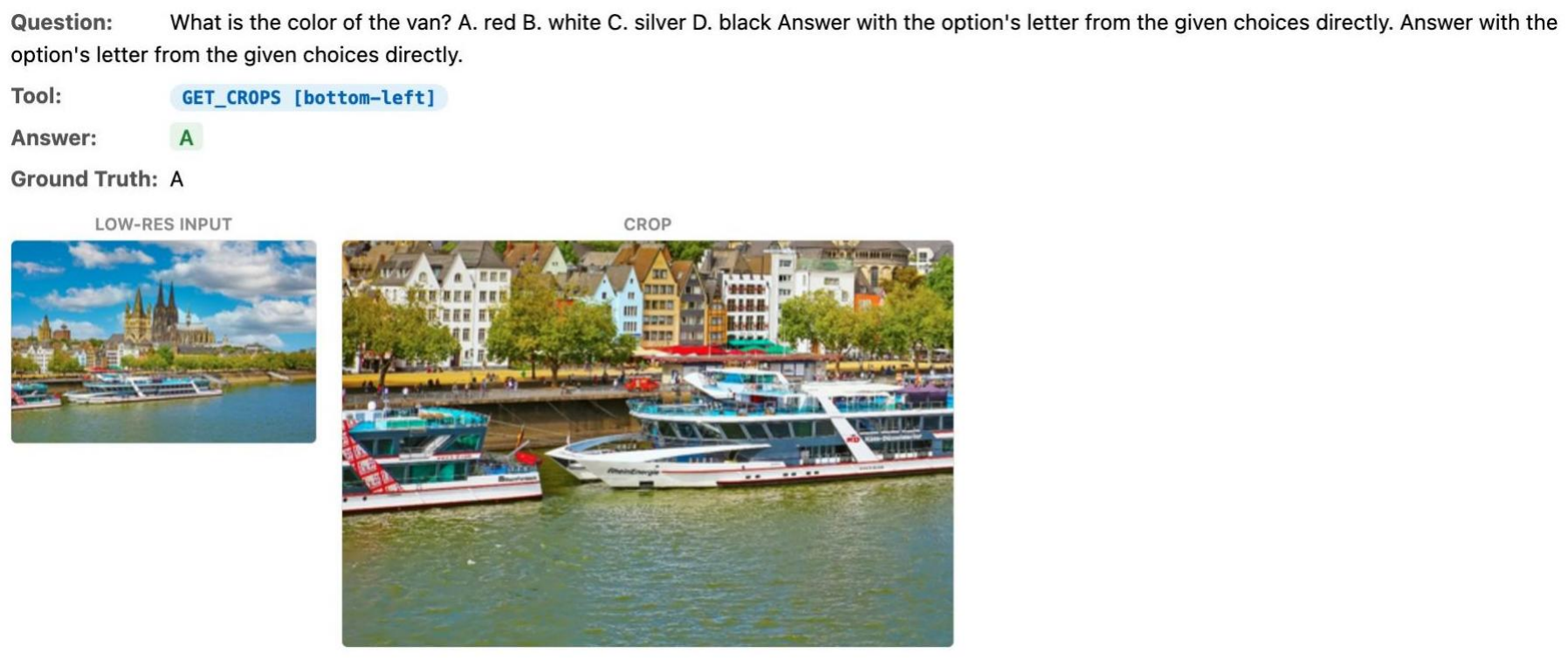} \\[1pt]
        \midrule \\[-4pt]
        \includegraphics[width=0.9\textwidth]{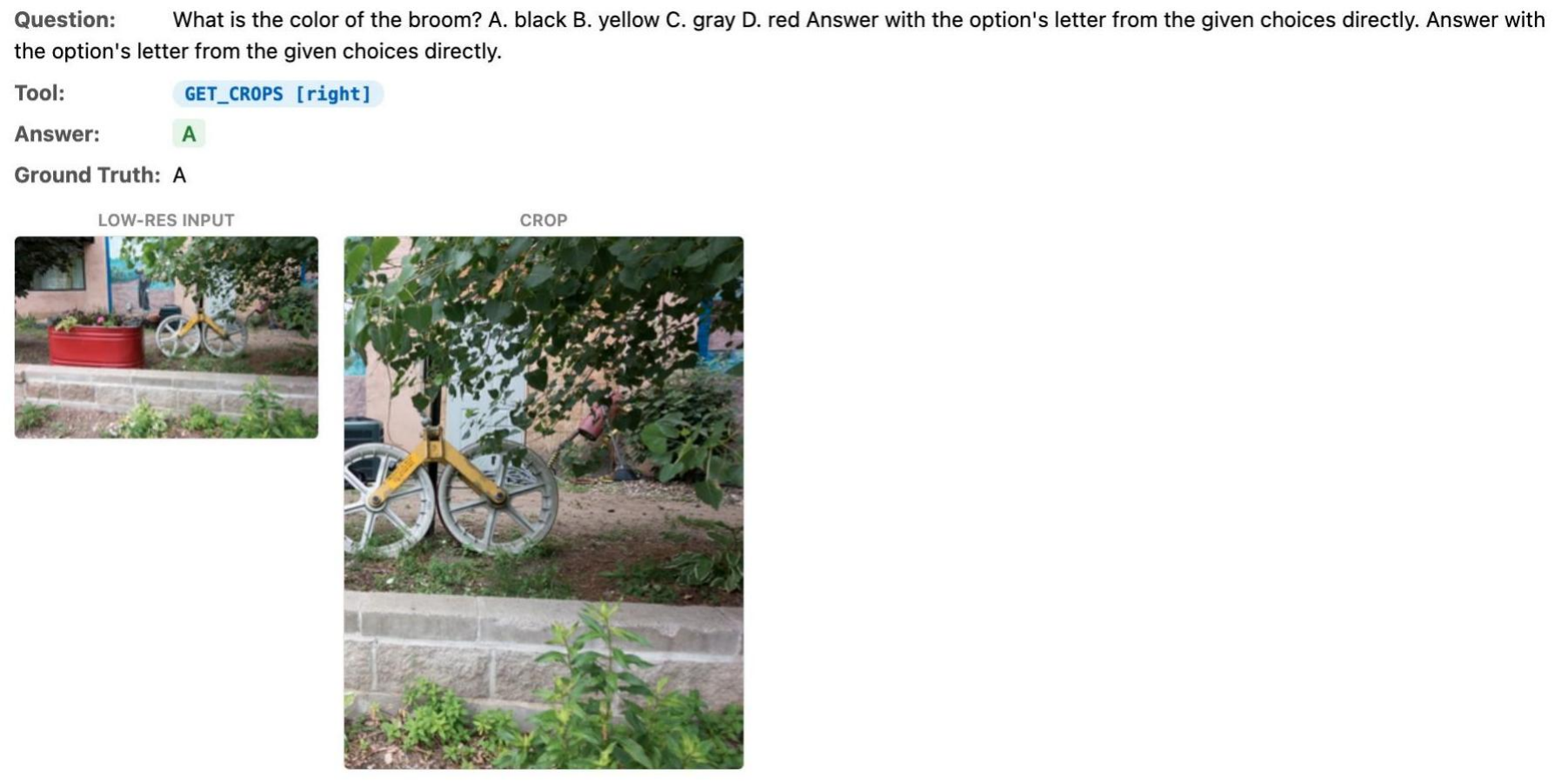} \\
    \end{tabular}
    \caption{\textbf{Positive examples of \method{}'s adaptive cropping (5/6).} The crops focuses on the church building top and detect the dove, zoom into the road next to the boats at a riverfront to resolve a van's color, and focus on a broom among garden clutter to identify its color. These examples show how \method{} localizes small or partially occluded objects that the question refers to, enabling fine-grained attribute recognition.}
    \label{fig:positive_examples_5}
\end{figure*}

\begin{figure*}[t]
    \centering
    \setlength{\tabcolsep}{1pt}
    \renewcommand{\arraystretch}{0.6}
    \begin{tabular}{c}
        \includegraphics[width=0.9\textwidth]{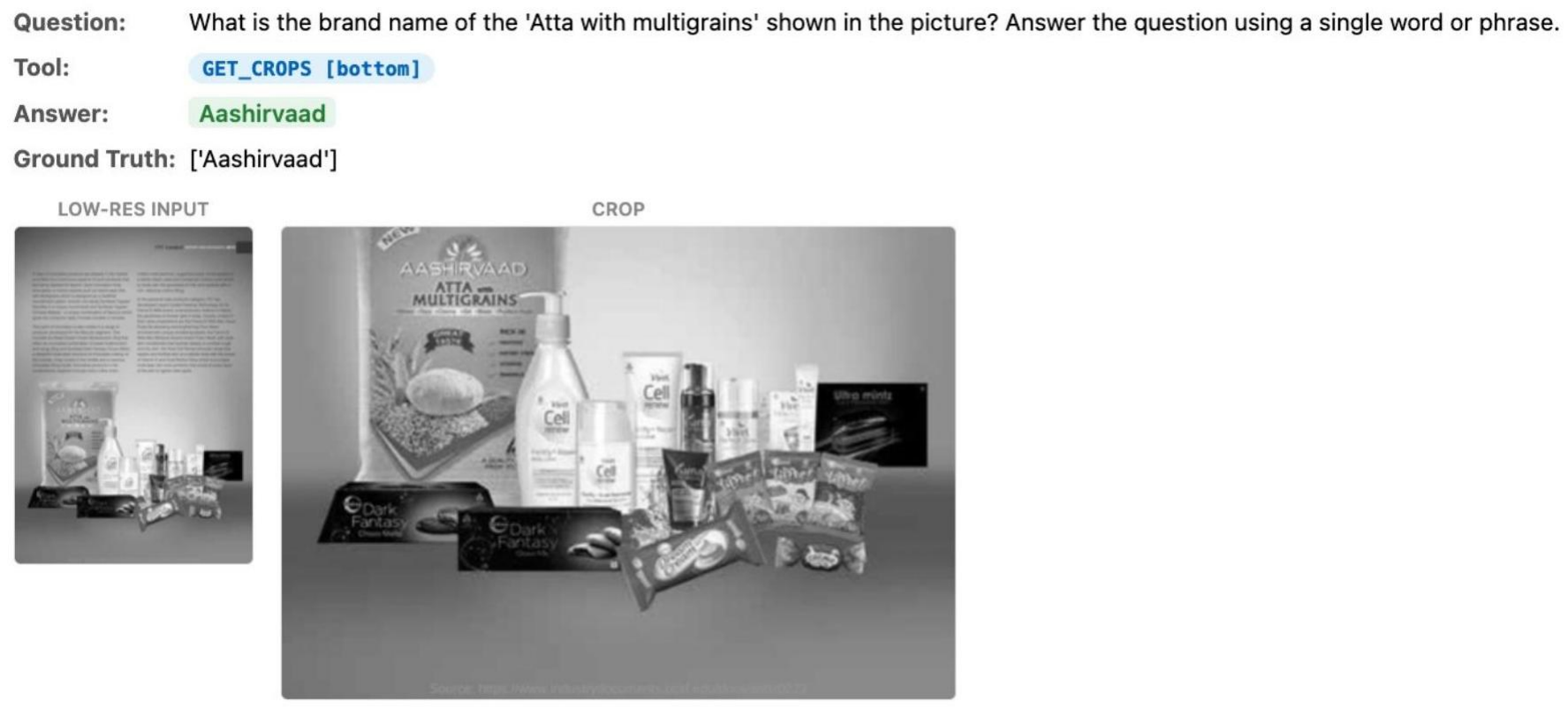} \\[1pt]
        \midrule \\[-4pt]
        \includegraphics[width=0.9\textwidth]{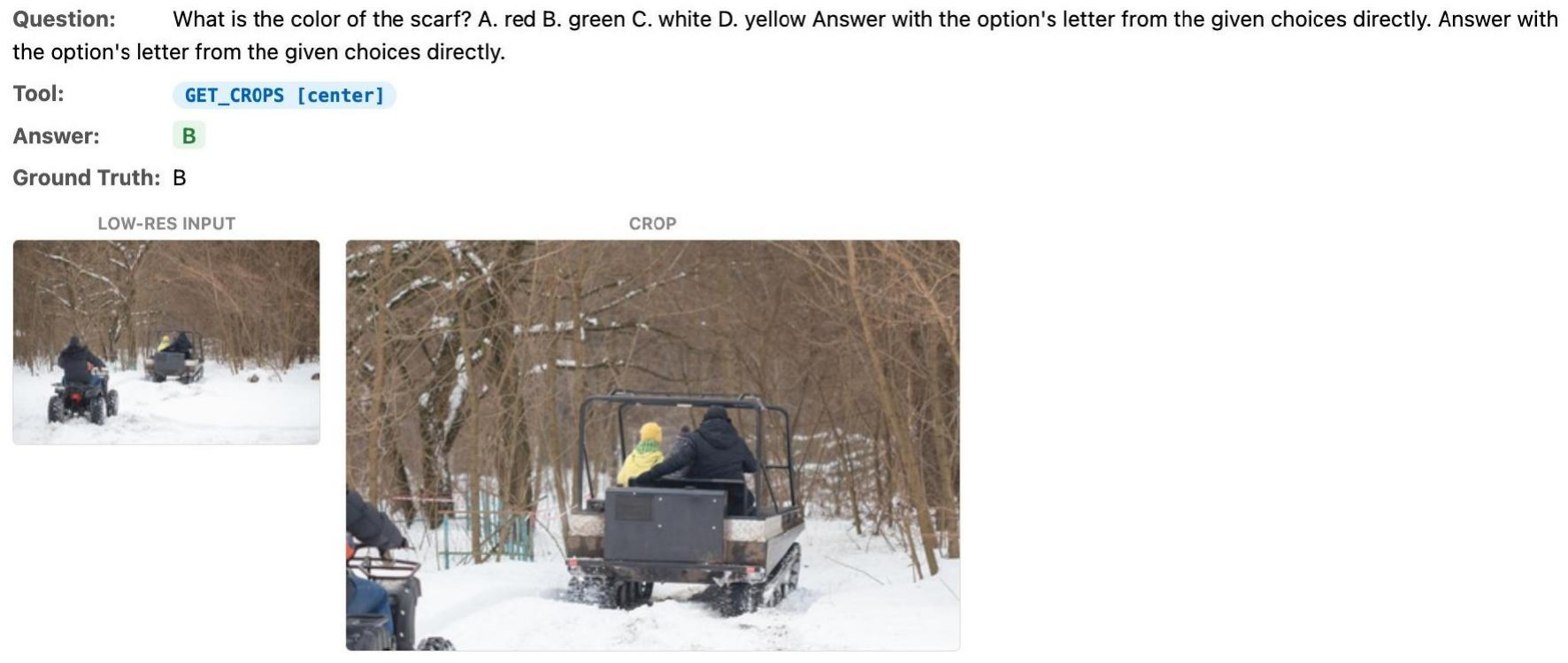} \\
    \end{tabular}
    \caption{\textbf{Positive examples of \method{}'s adaptive cropping (6/6).} The model focuses on product packaging in a cluttered display to identify a brand name, and crops into the center of a snowy scene to zoom in on a person riding an ATV, resolving the green color of their scarf.}
    \label{fig:positive_examples_6}
\end{figure*}

Figures~\ref{fig:positive_examples_1} - \ref{fig:positive_examples_6} present positive examples where \method{} successfully identifies and crops the image region relevant to the question. Each conversation shows the question, the tool call selected by the model, the predicted answer (matching the ground truth), the low-resolution input, and the retrieved high-resolution crop. Across charts, documents, and natural images, the retrieved high-resolution crops consistently isolate the task-relevant objects or text - such as axis labels, table cells, brand logos, and foreground subjects, enabling the model to extract fine-grained details that are unresolvable in the low-resolution overview alone.

\subsection{Failure cases}
\begin{figure*}[t]
    \centering
    \setlength{\tabcolsep}{1pt}
    \renewcommand{\arraystretch}{0.6}
    \begin{tabular}{c}
        \includegraphics[width=0.8\textwidth]{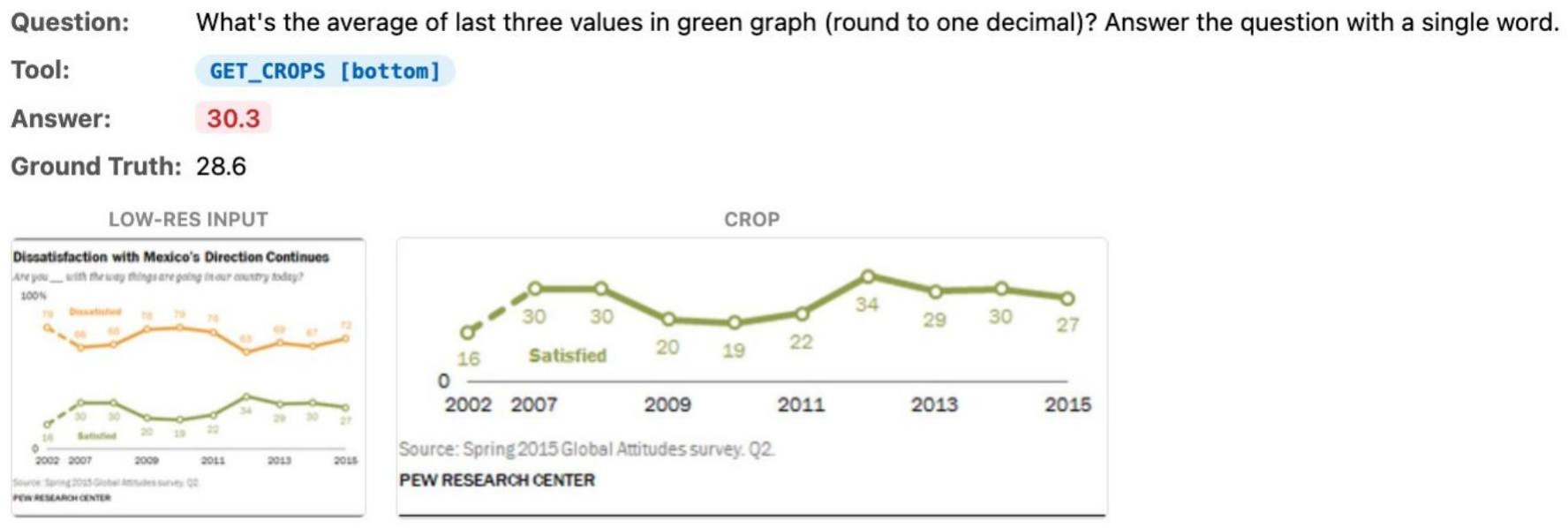} \\[1pt]
        \midrule \\[-4pt]
        \includegraphics[width=0.8\textwidth]{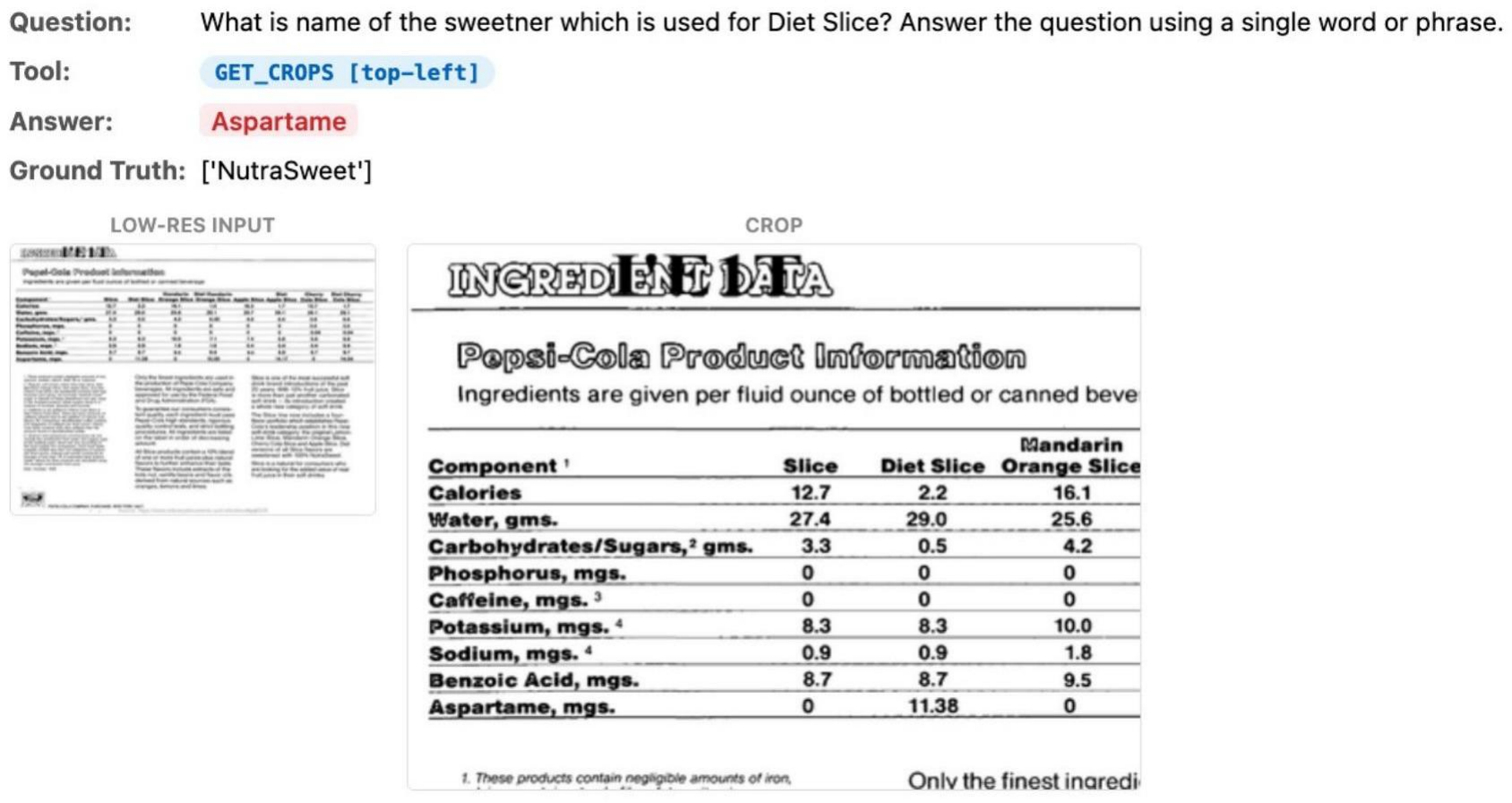} \\[1pt]
        \midrule \\[-4pt]
        \includegraphics[width=0.8\textwidth]{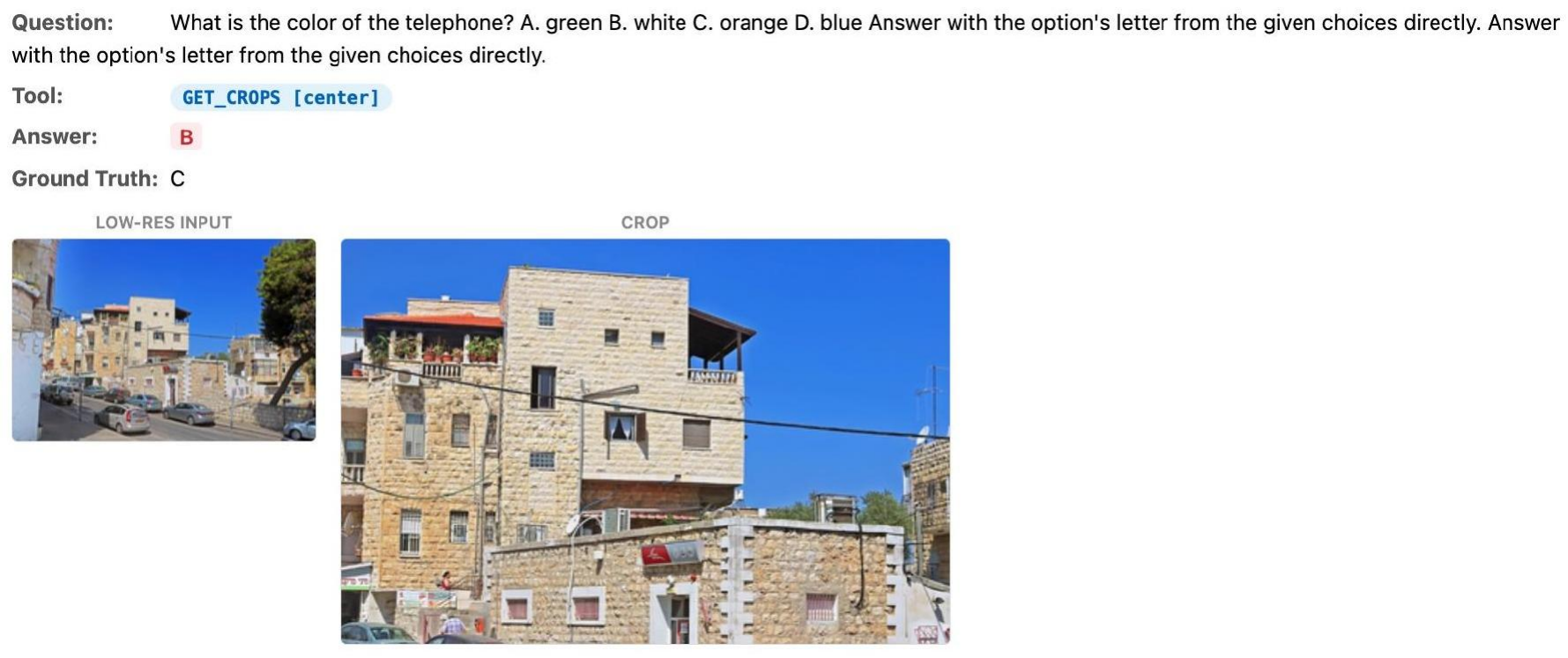} \\[1pt]
    \end{tabular}
    \caption{\textbf{Failure cases of \method{}'s adaptive cropping.} The crop either target the wrong region due to lack of sufficient context from the question, or produce a crop that targets a plausible region but fail to produce the correct answer. In the first example, the crop isolates the correct line, yet the model skips the value at 2013 and use 2012 instead, yielding the wrong average. In the second, the crop zooms into the \texttt{Nutrition Facts label}, instead of the ingredient list and reads ``Aspartame'' instead of the ``NutraSweet''. In the third, the crop focuses on the center, where the phone is located, but the granularity of the crop keep access of information around the object that makes the model misidentify its color.}
    \label{fig:negative_examples}
\end{figure*}

Figure~\ref{fig:negative_examples} shows representative failure cases where the model's either crop incorrect region (e.g. noisy annotations \cite{kimhi2025noisyannotationssemanticsegmentation,ryoo}), or correct crop that lead to a wrong answer. Each conversation shows the question, the tool call selected by the model, the predicted answer (in red) and the ground truth answer from the data, the low-resolution input, and the retrieved high-resolution crop.These failures typically arise when the question-relevant detail occupies a small or ambiguous portion of the image, causing the crop to capture nearby but insufficient context.

\section{Latency Analysis}
\label{sec:latency_details}
We analyze the latency by first describing the hardware and measurement setup, then reporting detailed wall-clock results, and finally comparing response lengths to explain the observed efficiency gains.

\subsection{Hardware and Measurement Setup}
To compare VisionThink~\cite{VisionThink} and \method{}, we evaluate both methods using their respective evaluation implementations in lmms-eval~\cite{lmmseval}, with both models running in native HuggingFace configuration. We measure wall-clock time (WC) from the start of generation until the final answer is produced (encompassing both turns when applicable).All measurements done on Nvidia-H100-80GB GPU.

\subsection{Detailed Latency Results}
Table~\ref{tab:latency_dynamic} presents the latency (WC) comparison between \method{} and competing dynamic methods across six benchmarks. \method{} achieves the lowest latency on every benchmark and the lowest average overall. On average, \method{} reduces wall-clock time by $4.4\times$ compared to VisionThink (from 2.71s to 0.61s), $3.9\times$ compared to PixelReasoner (2.37s with low-resolution first turn), and over $9\times$ compared to DeepEyes (6.03s-7.22s), whose iterative visual reasoning incurs the largest overhead. Even against the strongest baseline, ERGO with a low-resolution first turn (1.20s average), \method{} remains $2\times$ faster. The efficiency gains are most pronounced on memory- and reasoning-heavy benchmarks, where competing methods' extended reasoning traces or repeated high-resolution passes incur substantial overhead.

\begin{table*}
\centering
\caption{\textbf{Latency comparison of dynamic methods.} Per-benchmark latency (seconds) for crop-inspired and adaptive methods. ``LR'' indicates the first turn uses a low-resolution image. \method{} achieves the lowest average latency across all benchmarks. Best results per column in \textbf{bold}.}
\label{tab:latency_dynamic}
\resizebox{\textwidth}{!}{%
\begin{tabular}{l| c c c c c c | c}
\toprule
Model & ChartQA & DocVQA & OCRBench & POPE & RealWorld & V$^{*}$ Bench & \textbf{Average} \\
\midrule
ERGO          & 1.82 & 3.18 & 1.16 & 1.09  & 2.42 & 2.60 & 2.05 \\
ERGO (LR)     & 0.96 & 2.12 & 0.73 & 0.76 & 0.96 & 1.68 & 1.20 \\
DeepEyes      & 6.15 & 7.48 & 5.13 & 8.19  & 7.03 & 9.33 & 7.22 \\
DeepEyes (LR) & 6.36 & 5.43 & 4.51 & 4.83  & 6.79 & 8.26 & 6.03 \\
PixelReasoner      & 3.89 & 3.84 & 1.53 & 2.73  & 0.44 & 2.62 & 2.51 \\
PixelReasoner (LR) & 3.79 & 3.61 & 1.44 & 2.63  & 0.26 & 2.47 & 2.37 \\
VisionThink   & 4.32 & 1.78 & 3.36 & 1.23  & 2.31 & 3.24 & 2.71 \\
\midrule
\method{}     & \textbf{0.56} & \textbf{0.51} & \textbf{0.64} & \textbf{0.50} & \textbf{0.66} & \textbf{0.81} & \textbf{0.61} \\
\bottomrule
\end{tabular}
}
\end{table*}

\subsection{Response Length Comparison}
The latency differences stem primarily from response length. As shown in Table~\ref{tab:response_length}, VisionThink generates substantially longer reasoning traces, approximately $5.8\times$ to $28.8\times$ more characters than \method{}. This verbosity directly translates to increased generation time, as autoregressive decoding scales linearly with output length.

Beyond raw efficiency, \method{} exhibits significantly lower variance in response length across all benchmarks. This predictability has practical implications: estimated completion times become more reliable, enabling better resource allocation and user experience. In contrast, VisionThink's high standard deviations---often exceeding the mean---make response length and latency highly unpredictable, complicating deployment in latency-sensitive applications.

\begin{table}[t]
  \centering
  \caption{
  Response verbosity comparison between VisionThink and \method{} across the benchmarks. We report the mean $\pm$ std number of characters in model responses. \method{}
  produces substantially shorter responses than VisionThink ($5.8\times$--$28.8\times$ reduction), reflecting its efficiency beyond visual token savings.}
  \small
  \setlength{\tabcolsep}{3pt}
  \resizebox{\columnwidth}{!}{%
  \begin{tabular}{l r r r c}
  \toprule
  \textbf{Benchmark} & \textbf{samples} & \textbf{VisionThink} & \textbf{\method} & \textbf{Ratio} \\
  \midrule
  ChartQA       & 2500 & $89.46 \pm 205.71$  & $6.54 \pm 6.16$   & $13.7\times$ \\
  DocVQA (val)  & 5349 & $79.95 \pm 242.21$  & $13.74 \pm 12.79$ & $5.8\times$  \\
  OCRBench      & 1000 & $178.67 \pm 540.12$ & $16.23 \pm 16.39$ & $11.0\times$ \\
  POPE          & 9000 & $46.46 \pm 155.52$  & $2.84 \pm 2.33$   & $16.4\times$ \\
  RealWorldQA   &  765 & $160.67 \pm 261.55$ & $5.57 \pm 5.86$   & $28.8\times$ \\
  V*Bench       &  191 & $118.54 \pm 232.17$ & $5.29 \pm 5.78$   & $22.4\times$ \\
  \bottomrule
  \end{tabular}
  }
  \label{tab:response_length}
\end{table}


\section{Additional Cold-Start (SFT) Analysis}
\label{sec:supp_sft}

This supplementary section provides additional evidence that the cold-start stage learns a \emph{coupled-decision policy (CDP)} whose first-turn action jointly determines \emph{when} to request additional resolution ($C=\emptyset$ vs.\ $C\neq\emptyset$) and, when escalating, \emph{where} to look via the chosen crop subset $C\subseteq\mathcal{C}$.
Beyond final-task accuracy, we therefore report policy-centric diagnostics that quantify (i) the tendency to call the crop tool, (ii) failure to escalate when detail is required, and (iii) the amount of high-resolution evidence requested.

\subsection{Metrics for CDP diagnostics}
\label{subsec:supp_sft_metrics}

For each evaluated sample with resolution-sufficiency label $y\in\{\texttt{LR},\texttt{HR}\}$, the model either takes a no-call action ($C=\emptyset$) or requests one or more crops ($C\neq\emptyset$).
We evaluate the CDP as two coupled components:

\paragraph{Call decision (\emph{when} to crop).}
We treat tool invocation as a binary classifier where the positive class is $y=\texttt{HR}$ and a prediction is positive iff $C\neq\emptyset$.
We report:
\begin{itemize}
    \item \textbf{Call Precision} $\uparrow$: $\mathbb{P}(y=\texttt{HR}\mid C\neq\emptyset)$.
    \item \textbf{Call Recall} $\uparrow$: $\mathbb{P}(C\neq\emptyset\mid y=\texttt{HR})$.
    \item \textbf{Call F1} $\uparrow$: $2\,\mathrm{Prec}\times\mathrm{Rec}/(\mathrm{Prec}+\mathrm{Rec})$.
    \item \textbf{FPR (LR-call)} $\downarrow$: $\mathbb{P}(C\neq\emptyset\mid y=\texttt{LR})$.
\end{itemize}

\paragraph{Region decision (\emph{where} to look $\mid$ call).}
Conditioned on $C\neq\emptyset$, we measure overlap between the requested crop set and the grounding model target crops $\mathcal{C}^\star$ used for SFT supervision.
We report:
\begin{itemize}
    \item \textbf{Exact match} ($\mathrm{IoU}=1$) $\uparrow$: the predicted region matches the grounding model target exactly.
    \item \textbf{Relaxed match} ($\mathrm{IoU}\ge 0.25$) $\uparrow$: the predicted region overlaps a grounding model target by at least 0.25 IoU, accounting for the hierarchical structure of $\mathcal{C}$ (e.g., for a quadrant target, its two adjacent half-image regions are also considered acceptable; likewise, \texttt{All} and \texttt{Center} may be acceptable when they satisfy the IoU threshold).
    \item \textbf{Avg.\ area} $\downarrow$: $\mathbb{E}[s(C)]$, the average fraction of image area requested when calling.
\end{itemize}

In addition, we report \textbf{Accuracy} ($\uparrow$) and \textbf{RTR} ($\downarrow$) across all benchmarks, consistent with the main paper.

\subsection{CDP behavior across cold-start variants}
\label{subsec:supp_sft_cdp_diag}

Table~\ref{tab:sft_cdp_diag_pr} decomposes cold-start behavior into the two components of the coupled-decision policy (CDP): the \emph{call decision} (\emph{when} to request crops) and the \emph{region decision} (\emph{where} to look, conditioned on calling).

\paragraph{Call decision (\emph{when} to crop).}
Trajectory-level SFT substantially improves the calibration of the first-turn decision relative to baseline SFT.
While baseline SFT achieves moderate precision (62.49) it exhibits very low recall (15.34) and a high false-positive rate on LR samples (79.69), indicating that tool invocation is both unreliable on HR cases and overly frequent on LR cases.
Trajectory-level SFT increases recall to 41.02 and reduces FPR to 63.33, yielding a higher overall F1 score (24.63$\rightarrow$44.56).
Further upweighting the tool-call turn ($w_t{=}5$) strengthens this behavior: precision rises to 77.8, recall to 47.70, and FPR drops to 49.85, improving F1 to 59.14.
Together, these results indicate that emphasizing the first-turn action stabilizes the fused CDP decision of whether to escalate.

\paragraph{Region decision (\emph{where} to look $\mid$ call).}
Trajectory-level SFT also improves alignment with grounding model's supervision, increasing exact region match (IoU$=1$) from 13.8 to 15.9 and relaxed overlap match (IoU$\ge 0.25$) from 32.6 to 48.85, while reducing the average requested area from 0.59 to 0.402.
Upweighting the tool-call turn yields a large jump in localization quality (IoU$=1$: 41.3; IoU$\ge 0.25$: 75.5), with a modest increase in requested area (0.402$\rightarrow$0.463).
This suggests that stronger supervision of the tool-call turn improves not only the decision to escalate but also the selection of informative regions once escalation occurs.

Overall, improved cold-start recipes shape both parts of the CDP: they improve the reliability of escalation (\emph{when}) and the quality of evidence localization (\emph{where}), while keeping the requested high-resolution area controlled.
This motivates the subsequent GRPO stage, which further refines the same fused CDP under an explicit accuracy--efficiency objective.

\begin{table}[t]
\centering
\caption{\textbf{Cold-start (SFT) policy diagnostics for the coupled-decision policy (CDP).}
We evaluate the \emph{call decision} as a binary classifier where the positive class is $y=\texttt{HR}$ and a prediction is positive iff the model calls the tool ($C\neq\emptyset$).
We report precision, recall, and their harmonic mean (F1), as well as the false-positive rate on LR samples (FPR$=\mathbb{P}(C\neq\emptyset\mid y=\texttt{LR})$).
For the \emph{region decision} (conditioned on calling), we report overlap-based match rates to grounded targets using exact region match ($\mathrm{IoU}=1$) and relaxed overlap ($\mathrm{IoU}\ge 0.25$), along with the average requested area.
Higher is better ($\uparrow$) unless noted.}
\label{tab:sft_cdp_diag_pr}
\resizebox{\linewidth}{!}{%
\begin{tabular}{l | cccc | ccc}
\toprule
& \multicolumn{4}{c|}{\textbf{CDP: call decision}}
& \multicolumn{3}{c}{\textbf{CDP: region decision $\mid$ call}} \\
\cmidrule(lr){2-5}\cmidrule(lr){6-8}
Model
& Call Prec.$\uparrow$
& Call Rec.$\uparrow$
& Call F1$\uparrow$
& FPR (LR-call)$\downarrow$
& IoU=1$\uparrow$
& IoU$\ge$0.25$\uparrow$
& Avg. area$\downarrow$ \\
\midrule
SFT (baseline)
& 62.49 & 15.34 & 24.63 & 79.69
& 13.8 & 32.6  & 0.59 \\
SFT + Traj.
& 48.75 & 41.02 & 44.56 & 63.33
& 15.9 & 48.85  & 0.402 \\
SFT + Traj. + $w_t{=}5$
& 77.8 & 47.70 & 59.14 & 49.85
& 41.3 & 75.5 & 0.463 \\
\bottomrule
\end{tabular}}
\end{table}

\subsection{Cold-Start ablations}
We analyze how cold-start SFT design choices shape both downstream performance and the induced coupled-decision policy (CDP). We isolate three factors—trajectory-level supervision, phased (tool-first) scheduling, and high-resolution upsampling—and study their interaction with the tool-turn weight $w_t$. Beyond accuracy and RTR, we examine how each setting affects when and how aggressively the policy escalates, as well as the reliability of the tool-call interface itself.

\begin{table*}[t]
\centering
\small
\renewcommand{\arraystretch}{1.15}
\caption{Cold-start ablations. We report \textbf{Accuracy $\uparrow$} and \textbf{RTR $\downarrow$} across all benchmarks.}
\label{tab:supp_sft_ablation_full}
\resizebox{\textwidth}{!}{%
\begin{tabular}{ccc c | cc cc cc cc cc cc | cc}
\toprule
\multicolumn{3}{c}{\textbf{Components}} & & \multicolumn{2}{c}{ChartQA} & \multicolumn{2}{c}{DocVQA} & \multicolumn{2}{c}{OCRBench} & \multicolumn{2}{c}{POPE} & \multicolumn{2}{c}{RealWorld} & \multicolumn{2}{c}{V$^{*}$ Bench} & \multicolumn{2}{c}{\textbf{Average}} \\
\cmidrule(lr){1-3} \cmidrule(lr){5-6} \cmidrule(lr){7-8} \cmidrule(lr){9-10} \cmidrule(lr){11-12} \cmidrule(lr){13-14} \cmidrule(lr){15-16} \cmidrule(lr){17-18}
\textbf{Traj.} & \textbf{Phased} & \textbf{Ups.} & $w_t$ & Acc $\uparrow$ & RTR $\downarrow$ & Acc $\uparrow$ & RTR $\downarrow$ & Acc $\uparrow$ & RTR $\downarrow$ & Acc $\uparrow$ & RTR $\downarrow$ & Acc $\uparrow$ & RTR $\downarrow$ & Acc $\uparrow$ & RTR $\downarrow$ & Acc $\uparrow$ & RTR $\downarrow$ \\
\midrule
\no  & \no  & \no  & 1 & 72.9 & 0.35 & 93.1 & 0.42 & 73.7 & 0.32 & 85.8 & 0.29 & 64.1 & 0.33 & 61.3 & 0.42 & 75.15 & 0.36 \\
\yes & \no  & \no  & 1 & 69.8 & 0.38 & 93.8 & 0.32 & 76.9 & 0.54 & 87.8 & 0.30 & 70.5 & 0.55 & 68.6 & 0.48 & 77.90 & 0.43 \\
\yes & \no  & \yes & 1 & 76.9 & 0.30 & 93.2 & 0.27 & 72.9 & 0.41 & 84.7 & 0.25 & 69.0 & 0.43 & 66.0 & 0.55 & 77.12 & 0.37 \\
\yes & \yes & \no  & 1 & 70.9 & 0.36 & 93.4 & 0.40 & 76.1 & 0.32 & 86.6 & 0.30 & 68.8 & 0.34 & 64.4 & 0.41 & 76.70 & 0.36 \\
\yes & \yes & \yes & 1 & 70.8 & 0.33 & 92.7 & 0.37 & 72.7 & 0.31 & 85.7 & 0.28 & 66.8 & 0.33 & 63.9 & 0.43 & 75.43 & 0.33 \\
\midrule
\yes & \no  & \no  & 1 & 69.8 & 0.38 & 93.8 & 0.32 & 76.9 & 0.54 & 87.8 & 0.30 & \textbf{70.5} & 0.55 & 68.6 & 0.48 & 77.90 & 0.43 \\
\yes & \no  & \no  & 5 & \textbf{77.0} & 0.42 & \textbf{94.0} & 0.35 & \textbf{78.8} & 0.61 & \textbf{88.0} & 0.32 & 69.7 & 0.64 & \textbf{70.7} & 0.60 & \textbf{79.70} & 0.49 \\
\bottomrule
\end{tabular}}
\end{table*}

\subsection{Cold-start sensitivity to data and parameters}
\label{subsec:supp_sft_sensitivity}
Table~\ref{tab:sft_sensitivity} summarizes how common cold-start knobs shape both downstream performance and the induced coupled-decision policy (CDP).
Trajectory-level SFT improves average accuracy over vanilla (77.90 vs.\ 75.15), but it also shifts the policy toward more frequent crop requests (call rate 22.55\%$\rightarrow$25.21\%) while substantially reducing the average requested area (0.59$\rightarrow$0.402), indicating that the model learns to localize with smaller crops rather than relying on large regions.
Increasing the tool-turn weight to $w_t{=}5$ further strengthens the first-turn CDP action, yielding the best accuracy (79.70) but also the highest RTR (0.49), consistent with a policy that escalates more often (call rate 29.02\%) and requests slightly larger regions (Avg.\ area 0.463).

Beyond $w_t$, data and schedule choices can trade off calibration and efficiency.
In particular, HR upsampling is expected to reduce missed escalations by exposing the policy to more detail-critical instances, but may increase tool usage if applied aggressively; phased (tool-first) schedules can stabilize tool invocation but sometimes change how strongly the policy couples region selection to answer quality.
Overall, the best cold-start configuration is the one that yields a reliable CDP (low misses and good region selection) while keeping call rate and requested area controlled, leaving GRPO to fine-tune efficiency rather than compensating for frequent cold-start failures.

\begin{table}
\centering
\caption{\textbf{Sensitivity of cold-start SFT.}
We vary data/objective knobs and report both downstream performance (Avg.\ Accuracy and Avg.\ RTR across all benchmarks) and induced CDP behavior (call rate and average requested area, both measured over evaluation prompts).
Call rate is $\mathbb{P}(C\neq\emptyset)$ and Avg.\ area is $\mathbb{E}[s(C)]$ (fraction of image area requested when calling).}
\label{tab:sft_sensitivity}
\resizebox{\linewidth}{!}{%
\begin{tabular}{l | cccc | cccc}
\toprule
\textbf{Setting}
& HR upsample & $w_t$  & Traj. SFT & Phased
& Acc$\uparrow$ & RTR$\downarrow$ & Call rate & Avg. area$\downarrow$ \\
\midrule
Default SFT & \no & 1 & \no  & \no  & 75.15 & 0.36 & 22.55\% & 0.59 \\
Traj. only & \no & 1 & \yes & \no  & 77.90 & 0.43 & 25.21\% & 0.402 \\
Traj. + high $w_t$ & \no & 5 & \yes & \no  & 79.70 & 0.49 & 29.02\% & 0.463 \\
Traj. + HR upsample & \yes & 1 & \yes & \no  & 77.12 & 0.37 & 23.0\% & 0.35 \\
Phased (tool-first) & \no & 1 & \yes & \yes & 76.70 & 0.36 & 24.0\% & 0.44 \\
\bottomrule
\end{tabular}}
\end{table}

\subsection{Tool-call formatting reliability}
\label{subsec:supp_tool_format}

Finally, Table~\ref{tab:tool_format} quantifies tool-call formatting reliability after cold-start by measuring whether a generated tool call can be parsed into a valid crop subset $C\subseteq\mathcal{C}$ without any post-processing.
This isolates protocol learning from downstream accuracy: malformed outputs can prevent escalation even when the policy intends to call the tool, effectively breaking the CDP at the interface level.

Trajectory-level SFT substantially improves tool-call validity, reducing corruption from 10.17\% to 1.43\%.
Moreover, the remaining failures are not merely cosmetic: a large fraction of corrupted calls lead to \emph{incorrect} crop requests even after simple recovery (5.48\% for baseline SFT vs.\ 0.94\% for Traj.).
Upweighting the tool-call turn ($w_t{=}5$) eliminates formatting corruption entirely (100\% valid parse), supporting our design choice to explicitly emphasize the first-turn action during cold-start.

  \begin{table}
  \centering
  \small
  \renewcommand{\arraystretch}{1.15}
  \setlength{\tabcolsep}{6pt}
    \caption{\textbf{Tool-call formatting reliability after cold-start.}
    Valid parse indicates that the tool output can be parsed into a crop subset $C\subseteq\mathcal{C}$ without post-processing.
    Corrupt outputs are split into \emph{recoverable} cases (simple extraction of intended crop id(s) succeeds) and \emph{incorrect} cases (post-processing yields a wrong crop).}
  \label{tab:tool_format}
  \resizebox{\linewidth}{!}{%
  \begin{tabular}{l c ccc}
  \toprule
  & \multirow{2}{*}{Valid parse (\%)$\uparrow$} & \multicolumn{3}{c}{Corrupt outputs (\%)$\downarrow$} \\
  \cmidrule(lr){3-5}
  Model &  & Total & Recoverable & Incorrect \\
  \midrule
  SFT (baseline) & 89.83 & 10.17 & 4.69 & 5.48 \\
  SFT + Traj. & 98.57 & 1.43 & 0.49 & 0.94 \\
  SFT + Traj. + $w_t{=}5$ & 100.00 & 0.00 & 0.00 & 0.00 \\
  \bottomrule
  \end{tabular}}                  
  \end{table}

\section{GRPO ablations}
\label{sec:grpo_ablations}

We ablate the key design choices in our training pipeline: the role of the cold-start SFT stage, the structure of the tool-cost penalty, and the choice of correctness reward. Table~\ref{tab:supp_grpo_ablation} reports accuracy and retained token ratio (RTR) across six benchmarks.

\paragraph{Cold start vs.\ GRPO only.}
Training only the reference policy with SFT yields strong accuracy (79.70 average) but at a high compute cost (0.49 RTR), as the model has no incentive to crop economically. Conversely, running GRPO without a cold start drives RTR down sharply (0.31) but sacrifices accuracy (77.11), since the model never acquires a reliable cropping prior before reward optimization begins. \method\ combines both stages, retaining the accuracy of SFT (80.30) while approaching the efficiency of GRPO-only (0.36 RTR), confirming that the two stages are complementary rather than redundant.

\paragraph{Tool-cost penalty.}
The area-based cost is what couples accuracy to efficiency. Removing the tool cost entirely raises RTR to 0.51 with no accuracy benefit, as the model over-crops indiscriminately. Setting $\lambda=0$ retains the misuse penalty but drops the area term, leaving RTR at 0.42---better than no cost, but still well above \method's 0.36. This shows the area-based term, not just the binary misuse signal, is responsible for the efficiency gains.

\paragraph{Correctness reward.}
We compare our text-similarity reward (sentence-transformer) against ANLS and LLM-as-a-judge (LaaJ). All three reach comparable accuracy ($\sim$80) and efficiency ($\sim$0.37 RTR), but text similarity offers the best trade-off between semantic matching and robustness to short-answer phrasing variation, without the latency and cost of an LLM judge in the reward loop.
\begin{table*}[h!]
\centering
\caption{\textbf{GRPO Ablations.} Top: comparing training the model only for the $\pi_{\theta_{ref}}$ (SFT) vs only GRPO (no cold start). Middle: effect of the tool cost penalty, $\lambda=0$ removes the area-based cost while retaining the misuse penalty, and removing the tool cost entirely leads to further over-cropping. Bottom: comparison of correctness reward functions. Text similarity (sentence-transformer) balances semantic matching and robustness, outperforming both ANLS and LLM-as-a-judge on short-answer evaluation. RTR\ denotes the fraction of full-resolution compute used ($\downarrow$ is better).}
\label{tab:supp_grpo_ablation}
\resizebox{\textwidth}{!}{%
\begin{tabular}{l cc cc cc cc cc cc cc}
\toprule
& \multicolumn{2}{c}{ChartQA} & \multicolumn{2}{c}{DocVQA} & \multicolumn{2}{c}{OCRBench} & \multicolumn{2}{c}{POPE} & \multicolumn{2}{c}{RealWorldQA} & \multicolumn{2}{c}{V$^{*}$ Bench} & \multicolumn{2}{c}{Average} \\
\cmidrule(lr){2-3} \cmidrule(lr){4-5} \cmidrule(lr){6-7} \cmidrule(lr){8-9} \cmidrule(lr){10-11} \cmidrule(lr){12-13} \cmidrule(lr){14-15}
Model & Acc. $\uparrow$ & RTR $\downarrow$ & Acc. $\uparrow$ & RTR $\downarrow$ & Acc. $\uparrow$ & RTR $\downarrow$ & Acc. $\uparrow$ & RTR $\downarrow$ & Acc. $\uparrow$ & RTR $\downarrow$ & Acc. $\uparrow$ & RTR $\downarrow$ & Acc. $\uparrow$ & RTR $\downarrow$ \\
\midrule
SFT only  & 77.0 & 0.42 & 94.0 & 0.35 & 78.8 & 0.61 & 88.0 & 0.32 & 69.7 & 0.64 & 70.7 & 0.60 & 79.70 & 0.49 \\
GRPO only & 76.48 & 0.25 & 93.98 & 0.25 & 72.20 & 0.31 & 85.91 & 0.25 & 67.06 & 0.32 & 67.02 & 0.54 & 77.11 & 0.31 \\
\midrule
w/o tool cost & 80.60 & 0.44 & 94.60 & 0.37 & 81.50 & 0.64 & 87.80 & 0.34 & 69.40 & 0.65 & 71.20 & 0.62 & 80.85 & 0.51 \\
w/o area cost ($\lambda=0$) & 81.28 & 0.38 & 94.42 & 0.37 & 80.70 & 0.40 & 85.93 & 0.37 & 69.28 & 0.41 & 71.20 & 0.58 & 80.47 & 0.42 \\
\midrule
Text-Sim$\rightarrow$ ANLS & 79.84 & 0.29 & 94.35 & 0.25 & 80.00 & 0.41 & 85.78 & 0.23 & 69.67 & 0.43 & 70.16 & 0.57 & 79.96 & 0.37 \\
Text-Sim$\rightarrow$ LaaJ & 79.00 & 0.29 & 94.20 & 0.25 & 80.30 & 0.42 & 85.80 & 0.24 & 69.40 & 0.44 & 70.70 & 0.58 & 79.90 & 0.37 \\
\midrule
\midrule
\method & 80.64 & 0.32 & 94.43 & 0.28 & 81.30 & 0.42 & 85.73 & 0.27 & 68.5 & 0.43 & 71.20 & 0.41 & 80.30 & 0.36 \\
\bottomrule
\end{tabular}
}
\end{table*}

\subsection{GRPO learns beyond grounding annotations}
\label{subsec::oracle_no_distillation}

The grounding model provides only noisy localization supervision for cold-start SFT: it does not provide final answers. The final policy is optimized by GRPO with answer-correctness and crop-cost rewards, shifting from grounded model annotations towards better policy.

\begin{figure}[H]
\centering
\includegraphics[width=\linewidth]{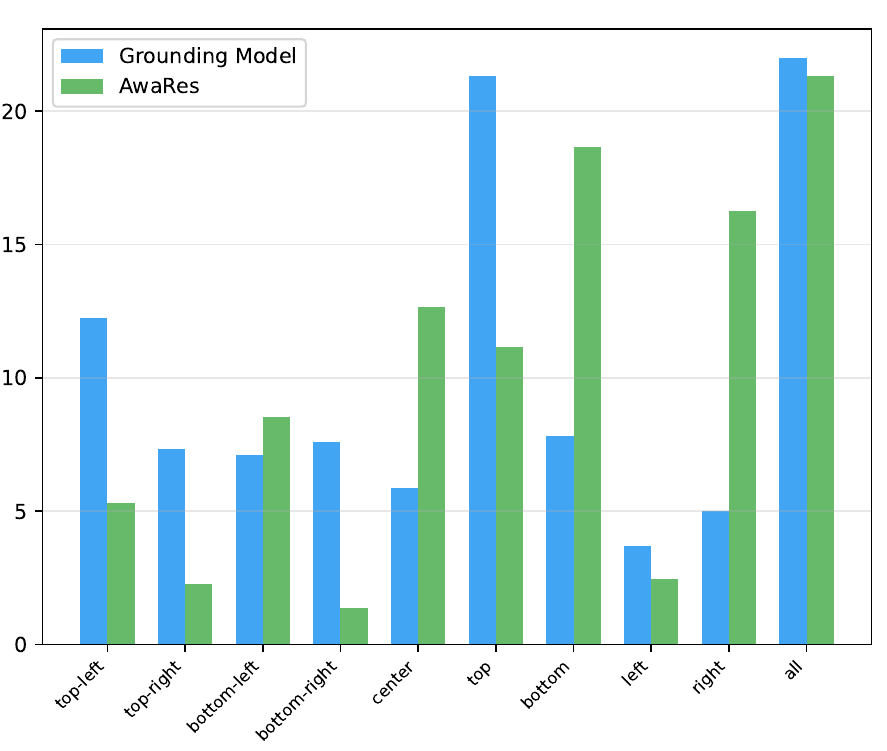}
\caption{GRPO changes the crop policy relative to grounding model supervision instead of copying it's crop choices.}
\label{fig:crop-dist}
\end{figure}

First, zero-shot grounding model tool use is insufficient. The strong grounding model drops from $88.9$ average accuracy in native HR mode to $79.89$ under the tool interface, with large drops on $V^\ast$ ($90.6\!\rightarrow\!73.8$), ChartQA ($90.8\!\rightarrow\!69.7$), and OCRBench ($86.9\!\rightarrow\!77.8$). AwaRes, despite being $7$B, reaches $80.30$ average accuracy, and outperforms the tool-prompted grounding model (MOE, 235B-A22B) on ChartQA ($80.6$ vs. $69.7$) and OCRBench ($81.3$ vs. $77.8$). Thus, grounding model annotations alone do not solve the task. The learned crop distribution differs from the grounding model annotations, as shown in Figure~\ref{fig:crop-dist}

\section{Training Details}
\label{sec:training_details}

We provide detailed hyperparameters for both training stages of \method{}: Cold-Start SFT and Tool Optimization via GRPO.
Table \ref{tab:training_params} conclude all the training parameters for both stages.

\begin{table}[t]
\centering
\caption{Training hyperparameters for Cold-Start SFT and Tool Optimization (GRPO) stages.}
\label{tab:training_params}
\small
\setlength{\tabcolsep}{3pt}
\resizebox{\columnwidth}{!}{%
\begin{tabular}{lcc}
\toprule
\textbf{Parameter} & \textbf{Cold-Start (SFT)} & \textbf{Tool Opt. (GRPO)} \\
\midrule
\multicolumn{3}{l}{\textit{Model Configuration}} \\
Base model & Instruct HF model & Cold-Start model \\
LORA Rank & 8 & 8 \\
\midrule
\multicolumn{3}{l}{\textit{Optimization}} \\
Optimizer & AdamW & AdamW \\
Learning rate & 1e-4 & $5\mathrm{e}{-5}$ \\
LR schedule & Cosine & Linear \\
Warmup ratio & 0.05 & 0 \\
Weight decay & 0.001 & 0 \\
Max gradient norm & - & 1.0 \\
\midrule
\multicolumn{3}{l}{\textit{Batch Configuration}} \\
Per-device batch size & 16 & 8 \\
Effective batch size & 128 & 64 \\
Number of epochs & 2 & 1 \\
\midrule
\multicolumn{3}{l}{\textit{Sequence Length}} \\
Max seq. / prompt length & 8192 & 512 \\
Max completion length & 512 & 512 \\
\midrule
\multicolumn{3}{l}{\textit{Stage-Specific}} \\
Tool-turn weight $w_t$ & 5 & -- \\
Trajectory-level opt. & True & -- \\
Number of generations $G$ & -- & 8 \\
Temperature $\tau$ & -- & 1.0 \\
Top-$p$ sampling & -- & 1.0 \\
KL coefficient $\beta$ & -- & 0.05 \\
Crop cost weight $(1-\alpha)$ & -- & 0.25 \\
Area weight $(1-\alpha)$ & -- & 0.01 \\
Answer reward weight $w_a$ & -- & 10.0 \\
\bottomrule
\end{tabular}%
}
\end{table}

\raggedbottom
\onecolumn
\section{Prompts}
\label{sec:prompts}
\subsection{LLM-as-a-Judge Prompt}
\begin{tcolorbox}[
    colback=gray!10,
    colframe=gray!50,
    boxrule=0.5pt,
    arc=2pt,
    left=6pt,
    right=6pt,
    top=6pt,
    bottom=6pt,
    title={\small\textbf{LLM-as-Judge Prompt}},
    coltitle=black,
    colbacktitle=gray!25
]
{\ttfamily\small
You are tasked with evaluating which of two AI responses better answers the given question based on the provided ground truth.

Original prompt: \{prompt\}

Ground Truth Answer: \{gt\}

Response 1:\\
\{lr\_resp\}

Response 2:\\
\{hr\_resp\}

Compare both responses against the ground truth answer and respond with just one number, DO NOT include any other text:\\
0 - Both responses are equally good or equally bad, no significant difference\\
1 - Response 1 is better\\
2 - Response 2 is better
}
\end{tcolorbox}

\subsection{Grounding Prompt}
\noindent
\begin{tcolorbox}[colback=gray!10, colframe=gray!50, boxrule=0.5pt, arc=2pt, left=6pt, right=6pt, top=6pt, bottom=6pt, title={\small\textbf{Grounding model Localization Prompt}}, coltitle=black, colbacktitle=gray!25]
{\ttfamily\small You are an expert visual assistant. Your task is to analyze an image and determine which regions are relevant for both understanding and answering a given question.

\textbf{Instructions:}\\
1. Analyze the question and identify what visual information is needed\\
2. Localize TWO types of regions:\\
\hspace*{1em}a) QUESTION region: The area containing visual elements mentioned in or relevant to the question\\
\hspace*{1em}b) ANSWER region: The area containing the answer to the question\\
3. For each region type, provide a bounding box in normalized coordinates [x\_min, y\_min, x\_max, y\_max]\\
\hspace*{1em}- Coordinates should be between 0 and 1000 (normalized to image dimensions)\\
\hspace*{1em}- x\_min, y\_min: top-left corner of the bounding box\\
\hspace*{1em}- x\_max, y\_max: bottom-right corner of the bounding box\\
4. Provide tight bounding boxes that closely encompass only the relevant text/visual elements\\
5. Output ONLY a Python dictionary with two keys: 'question' and 'answer' - NOTHING ELSE!!

\textbf{CRITICAL:} Your entire response must be ONLY a Python dictionary. Do not include any explanation, reasoning, or other text.

\textbf{Output format:}\\
\{\\
\hspace*{1em}'question': [[x\_min, y\_min, x\_max, y\_max]],\\
\hspace*{1em}'answer': [[x\_min, y\_min, x\_max, y\_max]]\\
\}

\textbf{Notes:}\\
- If the question and answer regions overlap significantly or are the same, you may provide the same coordinates for both\\
- If multiple regions are needed for either question or answer, include multiple bounding boxes in the list\\
- Keep bounding boxes as tight as possible around the relevant text or visual elements

Your response must be ONLY: \{'question': [[x\_min, y\_min, x\_max, y\_max], ...], 'answer': [[x\_min, y\_min, x\_max, y\_max], ...]\}\\
No other text, no explanation, just the dictionary.\\
Question: ...
}
\end{tcolorbox}

\subsection{SFT / GRPO / Inference Prompt}
\begin{tcolorbox}[
    colback=gray!10,
    colframe=gray!50,
    boxrule=0.5pt,
    arc=2pt,
    left=6pt,
    right=6pt,
    top=6pt,
    bottom=6pt
]
\texttt{\small You are a vision-language model that analyzes images and answers questions about them. If the image resolution is too low for accurate analysis, respond with \textbf{GET\_CROPS:} followed by a list of crop numbers in square brackets (e.g., GET\_CROPS: ['3'] or GET\_CROPS: ['0', '5']), where the available crop numbers and their corresponding areas are \{'0': 'top-left', '1': 'top-right', '2': 'bottom-left', '3': 'bottom-right', '4': 'center', '5': 'top', '6': 'bottom', '7': 'left', '8': 'right', 'all': 'all'\}, otherwise provide your answer.}
\end{tcolorbox}